\documentclass[lettersize,journal]{IEEEtran}
\usepackage{amsmath,amsfonts}
\usepackage{algorithmic}
\usepackage{algorithm}
\usepackage{array}
\usepackage{textcomp}
\usepackage{stfloats}
\usepackage{url}
\usepackage{verbatim}
\usepackage{graphicx}
\usepackage{cite}

\usepackage{booktabs}
\usepackage{multirow}
\usepackage{subfigure}

\usepackage{enumitem}
\usepackage[table]{xcolor}
\usepackage{balance}

\usepackage{hyperref}
\hypersetup{
	colorlinks=true,
	linkcolor=red,
	filecolor=red,      
	urlcolor=red,
	citecolor=green,
}

\begin{document}

\title{Integrating Listwise Ranking into Pairwise-based Image-Text Retrieval}

\author{
	Zheng Li, 
	Caili Guo,~\IEEEmembership{Senior Member,~IEEE}, 
	Xin Wang,
	Zerun Feng, 
	Yanjun Wang
	% <-this % stops a space
	\thanks{This work was supported in part by the Fundamental Research Funds for the Central Universities~(2021XD-A01-1), in part by the Key Program of National Natural Science Foundation of China~(No. 92067202), and in part by the BUPT Excellent Ph.D. Students Foundation~(CX2020104). \textit{(Corresponding author: Caili Guo.)}}
	\thanks{Zheng Li, Xin Wang and Zerun Feng are with the Beijing Key Laboratory of Network System Architecture and Convergence, School of Information and Communication Engineering, Beijing University of Posts and Telecommunications, Beijing 100876, China~(e-mail: lizhengzachary@bupt.edu.cn; wangxin1999@bupt.edu.cn; fengzerun@bupt.edu.cn).}
	\thanks{Caili Guo is with the Beijing Laboratory of Advanced Information Networks,  School of Information and Communication Engineering, Beijing University of Posts and Telecommunications, Beijing 100876, China~(e-mail: guocaili@bupt.edu.cn).}
	\thanks{Yanjun Wang is with China Telecom Digital Intelligence Technology Co., Ltd., Beijing 100035, China~(e-mail: wangyanjun1@chinatelecom.cn).}
}

%\markboth{Submitted to IEEE Transactions on Circuits and Systems for Video Technology}

\maketitle

\begin{abstract}
Image-Text Retrieval~(ITR) is essentially a ranking problem.
Given a query caption, the goal is to rank candidate images by relevance, from large to small.
The current ITR datasets are constructed in a pairwise manner.
Image-text pairs are annotated as positive or negative.
Correspondingly, ITR models mainly use pairwise losses, such as triplet loss, to learn to rank.
Pairwise-based ITR increases positive pair similarity while decreasing negative pair similarity indiscriminately.
However, the relevance between dissimilar negative pairs is different.
Pairwise annotations cannot reflect this difference in relevance.
In the current datasets, pairwise annotations miss many correlations.
There are many potential positive pairs among the pairs labeled as negative.
Pairwise-based ITR can only rank positive samples before negative samples, but cannot rank negative samples by relevance.
In this paper, we integrate listwise ranking into conventional pairwise-based ITR.
Listwise ranking optimizes the entire ranking list based on relevance scores.
Specifically, we first propose a Relevance Score Calculation~(RSC) module to calculate the relevance score of the entire ranked list.
Then we choose the ranking metric, Normalised Discounted Cumulative Gain~(NDCG), as the optimization objective.
We transform the non-differentiable NDCG into a differentiable listwise loss, named Smooth-NDCG~(S-NDCG).
Our listwise ranking approach can be plug-and-play integrated into current pairwise-based ITR models.
Experiments on ITR benchmarks show that integrating listwise ranking can improve the performance of current ITR models and provide more user-friendly retrieval results.
The code is available at \href{https://github.com/AAA-Zheng/Listwise_ITR}{https://github.com/AAA-Zheng/Listwise\_ITR}.
\end{abstract}


\IEEEpeerreviewmaketitle

\section{Introduction}
\IEEEPARstart{I}{mage-Text} Retrieval~(ITR) is essentially a ranking problem~\cite{zhang2020context, chen2020imram, qu2021dynamic}.
In the case of text-to-image retrieval, given a query caption, the goal is to rank candidate images by relevance, from large to small.
On the other hand, image-to-text retrieval starts with a query image, and the goal is to rank candidate captions by relevance.
The challenge of ITR is the heterogeneous gap between images and captions.
The mainstream approach to ITR is to learn a model to measure the similarities between images and captions~\cite{wei2023less, zhu2023esa, dong2022hierarchical, li2023integrating}.
Then, the model obtains retrieval results by ranking the similarities.

\begin{figure}[!t]
	\centering
	\includegraphics[width=\linewidth]{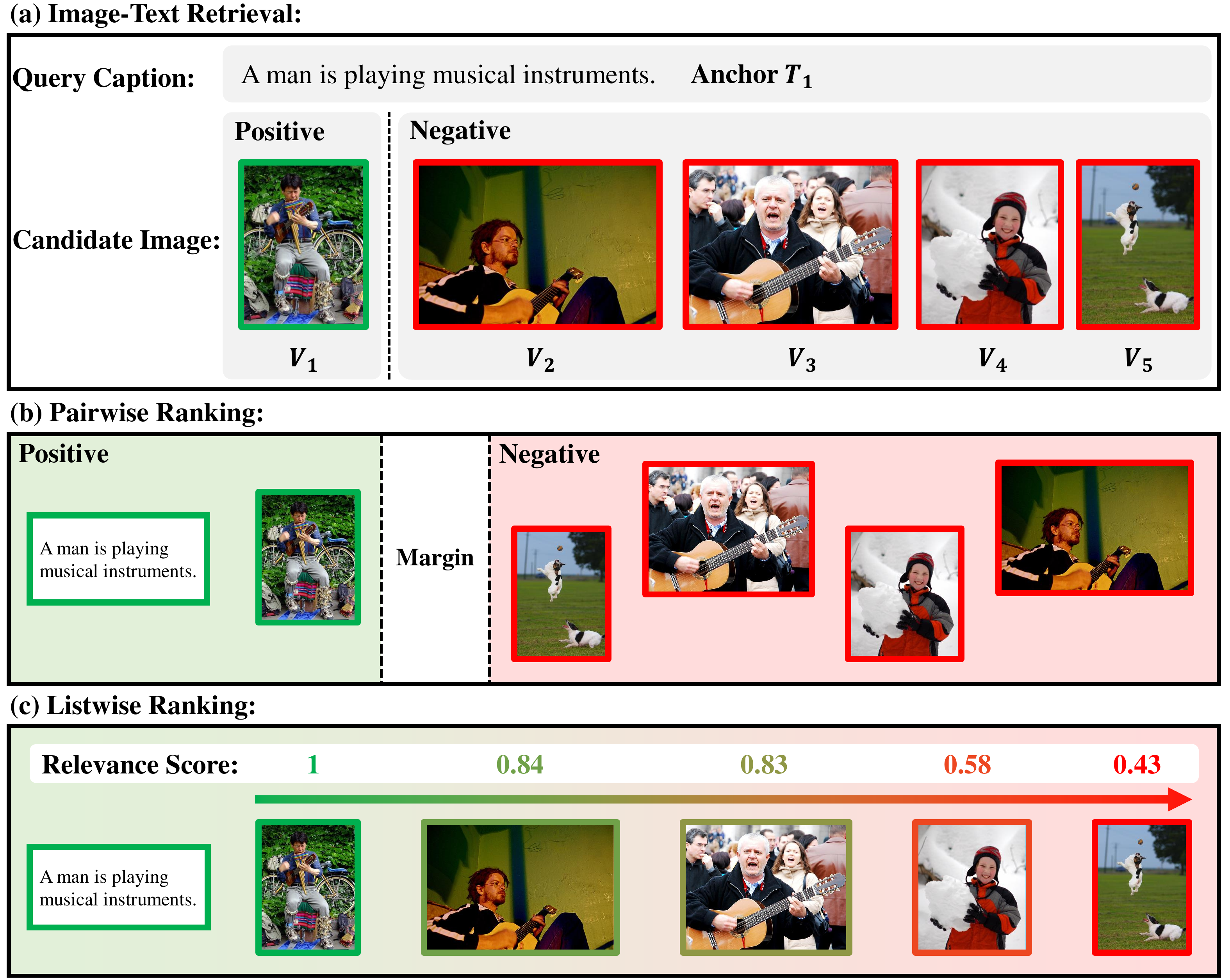}
	\caption{ITR based on pairwise and listwise ranking. The colors in~(c) indicate the relevance between image-text pairs. The closer the color is to green, the more relevant it is; the closer it is to red, the less relevant it is.}
	\label{example}
\end{figure}
The current ITR datasets are typically constructed in a pairwise manner.
Each image is annotated with several captions.
An image forms positive pairs with its annotated captions and negative pairs with other captions.
Correspondingly, the current ITR models mainly use pairwise losses to learn to rank.
Pairwise-based ITR increases positive pair similarity while decreasing negative pair similarity.
Triplet loss~\cite{schroff2015facenet} is one of the most commonly used pairwise losses.
A number of state-of-the-art ITR models~\cite{lee2018stacked, li2019visual, liu2020graph, diao2021similarity, zhang2022negative} employ triplet loss as the optimization objective.
A triplet consists of three parts: an anchor sample, a positive sample, and a negative sample.
As shown in \figurename~\ref{example}(a), taking text-to-image retrieval as an example, given a query caption $ T_{1} $, the goal is to rank candidate images by relevance, from large to small.
We take the query caption $ T_{1} $ as an anchor.
$ T_{1} $ is an annotation of image $ V_{1} $.
$ T_{1} $ and $ V_{1} $ form a positive pair.
$ T_{1} $ forms negative pairs with other images, such as $ \{V_{2}, V_{3}, V_{4}, V_{5}\} $.
Triplet loss aims to make the similarity of positive pairs larger than that of negative pairs.
\figurename~\ref{example}(b) is a schematic diagram of pairwise ranking using triplet loss as the optimization objective.
Triplet loss forces positive similarities to be higher than negative ones by a margin, which guarantees model generalization.
Current studies on ITR optimization mainly focus on improving the pairwise loss~\cite{faghri2018vse++, chen2020adaptive, wei2020universal, wei2021universal, wei2021meta}.
Faghri~\textit{et al.}~\cite{faghri2018vse++} propose triplet loss with hard negative mining, which incorporates hard negatives in the triplet loss.
Wei~\textit{et al.}~\cite{wei2020universal} propose a universal weighting framework that analyzes various pairwise losses.
These improved pairwise ranking approaches further enhance discrimination and generalization.

Although the current pairwise-based ITR achieves promising retrieval performance, it only focuses on the position of positive samples in the ranking list.
Regardless of their relevance to the anchor, all negative samples are indiscriminately far from it.
However, the relevance between dissimilar negative pairs is different.
Pairwise annotations in current datasets can not reflect this difference in relevance.
As shown in \figurename~\ref{example}(a), $ \{V_{2}, V_{3}, V_{4}, V_{5}\} $ are all negative samples, but the relevance between them and the anchor is different.
From $ V_{2} $ to $ V_{5} $, the relevance between these negative samples and the anchor decreases in turn.
An ideal ITR model should rank candidates from high to low relevance, which cannot be achieved with pairwise ranking.
As \figurename~\ref{example}(b) shows, pairwise ranking ranks the positive image $ V_{1} $ before all negative images $ \{V_{2}, V_{3}, V_{4}, V_{5}\} $.
But these negative samples cannot be ranked correctly.
It is worth noting that although $ V_{2} $ and $ V_{3} $ are negative samples, they are highly correlated with anchors.
They are potentially positive samples that are labeled as negative, and they are called \textit{false negatives} in~\cite{chun2022eccv}.
Previous studies have shown that there are a large number of false negatives in existing ITR datasets~\cite{chun2022eccv}.
But the current pairwise ranking ignores the ranking of false negatives, which can negatively impact the user experience of the ITR system.

In this paper, we integrate listwise ranking into conventional pairwise-based ITR.
Listwise ranking optimizes the entire ranking list based on relevance scores.
Therefore, listwise ranking can rank all candidate samples by relevance, whether they are positive or negative samples.
There are two challenges to integrating listwise ranking into pairwise-based ITR:
\begin{itemize}[leftmargin=*]
	\item How to obtain the relevance score?
	\item How to optimize the entire ranking list?
\end{itemize}

The current ITR dataset only annotates positive and negative pairs.
The relevance scores between anchor and negative samples are not annotated.
To address this challenge, we propose a Relevance Score Calculation~(RSC) module. 
The RSC module approximates the relevance score by calculating the text similarity between captions.
The relevance score in \figurename~\ref{example}(c) is calculated by our RSC module.
It can be seen that the calculated relevance score is close to human judgment.

Regarding the second challenge, we optimize the entire ranked list by directly optimizing the ranking metric.
In the field of image retrieval, there are some studies~\cite{cakir2019deep, brown2020smooth} that train models by directly optimizing Average Precision~(AP).
However, AP can only measure the performance of rankings with binary relevance (positive or negative).
AP is insufficient for evaluating rankings with complex real-valued relevance scores.
In this paper, we choose the ranking metric, Normalised Discounted Cumulative Gain~(NDCG), as the optimization objective.
NDCG incorporates the relevance score of the ranking list into the evaluation metric.
Therefore, NDCG can comprehensively measure ranking quality, and is widely used in the evaluation of information retrieval and recommendation systems.
The difficulty with directly optimizing NDCG is that it is non-differentiable and cannot be optimized using gradient descent.
To this end, we transform the non-differentiable NDCG into a differentiable listwise loss, named Smooth-NDCG~(S-NDCG).
As a result, listwise ranking can be optimized using gradient descent.
\figurename~\ref{example}(c) shows the optimization process of listwise ranking.
Listwise ranking can not only rank positive samples before negative samples, but also correctly rank negative samples according to their relevance scores.

Training a deep learning model requires dividing the data into batches.
There is a gap between the NDCG on each batch and the NDCG on the entire dataset.
Optimizing S-NDCG within a batch does not generalize well to the entire dataset.
Pairwise losses, such as triplet loss, generally have better generalization due to the introduced margin.
In order to make up for the shortcoming of S-NDCG in generalization, we jointly use triplet loss and S-NDCG as the optimization objective.
The RSC module and S-NDCG loss constitute our listwise ranking approach, which is independent of the current pairwise-based ITR model.
Therefore, listwise ranking can be plug-and-play integrated into conventional pairwise-based ITR models to obtain more user-friendly retrieval results.	
The major contributions of this paper are summarized as follows:
\begin{itemize}[leftmargin=*]
	\item We integrate listwise ranking into conventional pairwise-based ITR for relevance ranking.
	To the best of our knowledge, we are the first to apply listwise ranking to ITR.
	\item We propose a Relevance Score Calculation~(RSC) module to calculate the relevance scores of the entire ranked list, which enables listwise ranking to be applied in ITR.
	\item We transform the non-differentiable ranking metric NDCG into a differentiable listwise loss, named Smooth-NDCG~(S-NDCG).
	As a result, listwise ranking can be optimized using gradient descent.
	\item Our listwise ranking approach can be plug-and-play integrated into pairwise-based ITR models.
	Experiments on ITR benchmarks show that integrating listwise ranking can improve the performance of current ITR models and provide more user-friendly retrieval results.
\end{itemize}

\section{Related Work}
\subsection{Image-Text Retrieval}
Image-Text Retrieval~(ITR) is essentially a ranking problem.
The mainstream approach to ITR is to learn a model to measure the similarities between images and captions.
Then, the model obtains retrieval results by ranking the similarities.
Existing ITR models can be mainly divided into two categories: global-level alignment methods and local-level alignment methods.

\subsubsection{Global-Level Alignment}
Global-level alignment methods embed the whole images and captions into a joint embedding space~\cite{li2019visual, wang2020consensus, chen2021learning, li2022multi}.
The similarity between the embeddings can be calculated using a simple similarity metric, \textit{e.g.}, cosine similarity.
Frome~\textit{et al.}~\cite{frome2013devise} propose the first global-level alignment model, DeViSE, which employs the Convolutional Neural Network~(CNN) and Skip-Gram to project images and captions into a joint embedding space.
Faghri~\textit{et al.}~\cite{faghri2018vse++} propose VSE++, which uses CNN to encode images and Gated Recurrent Unit~(GRU) to encode captions.
VSE++ incorporates hard negatives into triplet loss, which yields significant gains in retrieval performance.
Li~\textit{et al.}~\cite{li2019visual} propose a Visual Semantic Reasoning Network~(VSRN), which establishes connections between image regions.
VSRN uses the Graph Convolutional Network~(GCN) to generate global features with regional semantic associations.
Chen~\textit{et al.}~\cite{chen2021learning} propose a VSE$ \infty $ model, which introduces a Generalized Pooling Operator~(GPO) that learns the best pooling strategy for different features to generate global features.

\subsubsection{Local-Level Alignment}
Local-level alignment methods align image regions and words~\cite{lee2018stacked, chen2020imram, liu2020graph, zhang2020context, diao2021similarity}.
The similarities between image-text pairs are obtained by calculating the cross-attention between the region and words.
Lee~\textit{et al.}~\cite{lee2018stacked} propose a Stacked Cross Attention Network~(SCAN), which measures the similarity by aligning image regions and words.
Liu~\textit{et al.}~\cite{liu2020graph} propose a Graph Structured Matching Network~(GSMN) to learn fine-grained correspondence between images and captions.
Diao~\textit{et al.}~\cite{diao2021similarity} propose a SGRAF model, which contains a Similarity Graph Reasoning~(SGR) and Similarity Attention Filtration~(SAF) network for local and global alignments.

Our proposed listwise ranking approach is independent of the current ITR model architecture, and can plug-and-play to improve the retrieval performance of these models.

\subsection{Pairwise Ranking}
The current ITR datasets are constructed in a pairwise manner.
Image-text pairs are annotated as positive and negative pairs.
Correspondingly, the current ITR models mainly use pairwise losses to learn to rank.
In pairwise ranking, the relative order between two samples is taken into account.
In the ITR task, the current pairwise ranking guides the model to make positive pairs more similar than negative pairs, to achieve correct ranking.
Triplet loss~\cite{schroff2015facenet} is one of the most commonly used pairwise losses.
In ITR, a hinge-based triplet loss~\cite{frome2013devise} is widely used as an objective to force positive pairs to have higher similarities than negative pairs by a margin, which guarantees model generalization.
Faghri~\textit{et al.}~\cite{faghri2018vse++} propose triplet loss with hard negative mining, which incorporates hard negatives in the triplet loss and yields significant gains in retrieval performance.
Most state-of-the-art ITR models~\cite{lee2018stacked, li2019visual, liu2020graph, diao2021similarity, zhang2022negative} employ triplet loss as the optimization objective.
Recently, some studies propose more complex sample mining~\cite{chen2020adaptive} or sample weighting~\cite{wei2020universal, wei2021universal, wei2021meta} strategies, which further improve the retrieval performance of ITR models.
Although these pairwise ranking approaches achieves promising retrieval performance, it only focuses on the position of positive samples in the ranking list.
Regardless of their relevance to the anchor, all negative samples are indiscriminately far from it.
This can negatively impact the user experience of the ITR system.

\subsection{Listwise Ranking}
Listwise ranking optimizes the entire ranking list based on the relevance score.
The listwise loss is generally related to evaluation measures (\textit{e.g.}, an approximation or upper bound of the measure-based ranking errors).
There are several studies~\cite{chakrabarti2008structured, qin2010general, bruch2019revisiting} in the field of information retrieval dedicated to the direct optimization of ranking metrics such as AP and NDCG. 
These methods make remarkable progress in the fields of information retrieval and recommendation.
But they are not applied to the end-to-end training of deep neural networks.
Recently, some methods~\cite{cakir2019deep, brown2020smooth} for directly optimizing AP for image retrieval have been proposed.
FastAP~\cite{cakir2019deep} attempts to optimize AP within each batch using a soft histogram binning technique.
Brown~\textit{et al.}~\cite{brown2020smooth} introduce an objective that optimizes instead a smoothed approximation of AP, called Smooth-AP.
The direct optimization of evaluation metrics has been proven to improve training efficiency and performance~\cite{brown2020smooth}.
These AP optimization approachs are all applied to unimodal image retrieval.
In image retrieval data, the class to which each image belongs is unambiguous, and the relevance between images is binary (related or unrelated).
In ITR data, the relevance between pairs of negative samples is beyond binary.
The relevance between dissimilar negative samples and queries is different.
AP can only measure the performance of rankings with binary relevance scores and is insufficient for evaluating such rankings with complex real-valued relevance scores.
The NDCG measures the gain of candidates based on their position in the ranked list.
The gain is measured based on relevance scores between the query and candidates.
Therefore, we choose NDCG as the optimization objective and transform the non-differentiable NDCG into a differentiable listwise loss.

\section{Methodology}
\subsection{Preliminaries}
\subsubsection{Image-Text Retrieval}
Image-Text Retrieval~(ITR) is defined as retrieving relevant samples across images and captions.
We denote $ \mathcal{V} = \{ V_{i} \}_{i=1}^{I} $ as a collection of images and $ \mathcal{T} = \{ T_{i} \}_{i=1}^{I} $ as a collection of captions, where $ V_{i} $ is an image and $ T_{i} $ is a caption.
In the case of image-to-text retrieval, given an image $ V_{i} $, the goal is to find the most relevant caption $ T_{i} $ from the text collection $ \mathcal{T} $.
On the other hand, text-to-image retrieval starts with a caption $ T_{i} $, and the goal is to find the most relevant image $ V_{i} $ from the image collection $ \mathcal{V} $.
We denote a positive pair by $ (V_{i}, T_{i}) $ and a negative pair by $ (V_{i}, T_{j, i \neq j}) $.
Current ITR models generally follow a common framework, as shown in \figurename~\ref{framework}(a).
The framework contains an image encoder $ \boldsymbol{f}_{img}(\cdot) $ and a text encoder $ \boldsymbol{f}_{text}(\cdot) $.
The similarity $ s(V_{i}, T_{j}) $ is computed based on the encoded image feature $ \boldsymbol{f}_{img}(V_{i}) $ and text feature $ \boldsymbol{f}_{text}(T_{j}) $.
To simplify the representation, we abbreviate $ s(V_{i}, T_{j}) $ as $ s_{i,j} $ in the following.
Finally, the retrieval results are obtained by ranking the similarities.

\subsubsection{Pairwise Ranking}
\begin{figure}[t]
	\centering
	\includegraphics[width=\linewidth]{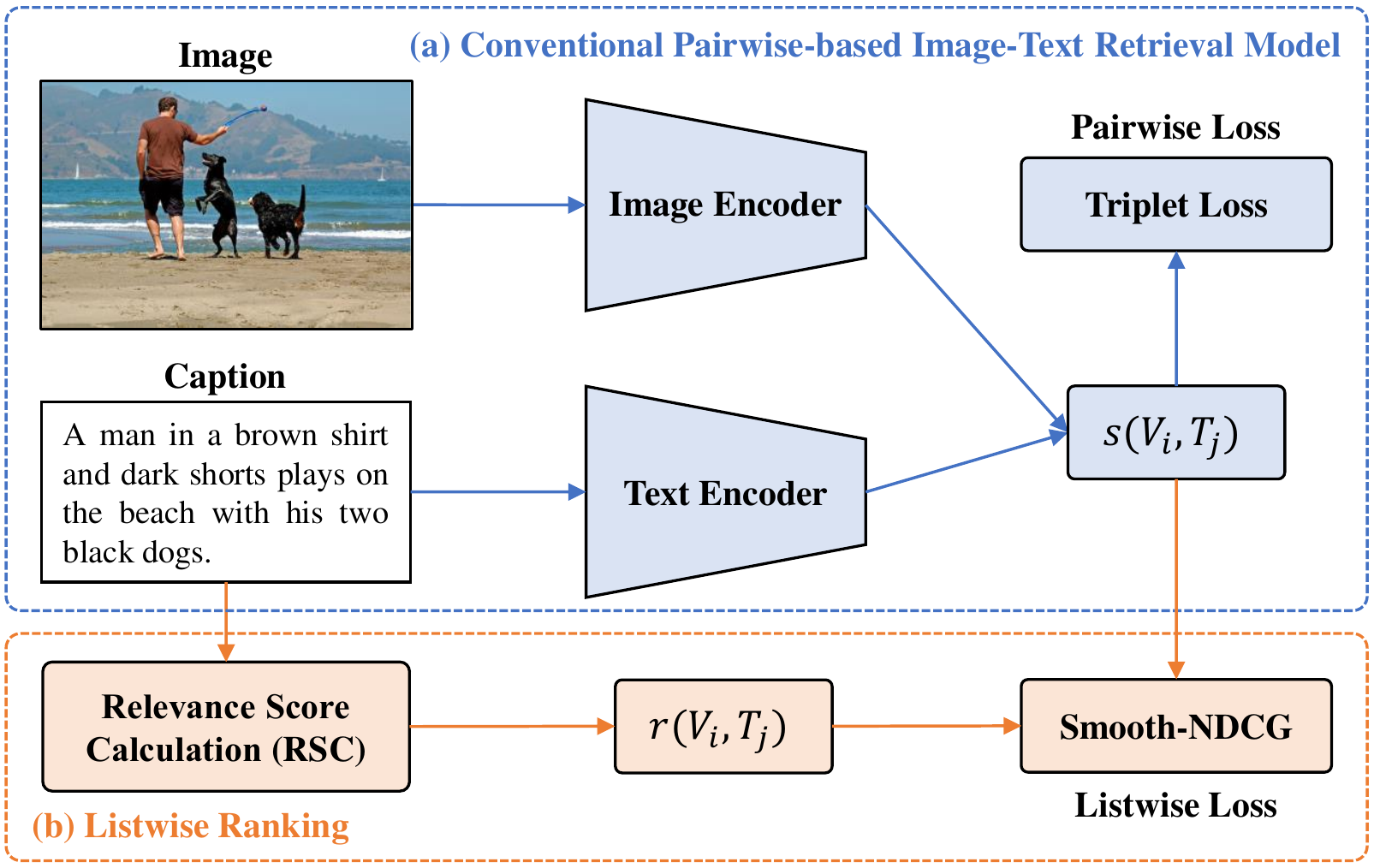}
	\caption{Image-text retrieval framework integrating listwise ranking.}
	\label{framework}
\end{figure}

The current ITR models mainly use pairwise losses to learn to rank.
Pairwise ranking guides the model to make positive pairs more similar than negative pairs to achieve the correct ranking.
Triplet loss~\cite{schroff2015facenet} is one of the most commonly used pairwise losses.
Most state-of-the-art ITR models~\cite{lee2018stacked, li2019visual, liu2020graph, diao2021similarity, zhang2022negative} employ triplet loss with hard negative mining as the optimization objective, which includes image-to-text~($ v \rightarrow t $) and text-to-image~($ t \rightarrow v $) directions:
\begin{equation}
	\mathcal{L}_{\text{Triplet}}
	= \mathcal{L}_{\text{Triplet}}^{v \rightarrow t}
	+ \mathcal{L}_{\text{Triplet}}^{t \rightarrow v}.
\end{equation}
$ \mathcal{L}_{\text{Triplet}}^{v \rightarrow t} $ takes the form of:
\begin{equation} \label{triplet}
	\mathcal{L}_{\text{Triplet}}^{v \rightarrow t}
	= \dfrac{1}{N}
	\sum_{i=1}^{N}
	\max_{j = 1, i \neq j}^{N}
	\left[
	s_{i,j} - s_{i,i} + \lambda 
	\right]_{+},
\end{equation}
$\mathcal{L}_{\text{Triplet}}^{t \rightarrow v} $ takes the form of:
\begin{equation}
	\mathcal{L}_{\text{Triplet}}^{t \rightarrow v}
	= \dfrac{1}{N}
	\sum_{i=1}^{N}
	\max_{j = 1, i \neq j}^{N}
	\left[
	s_{j,i} - s_{i,i} + \lambda
	\right]_{+},
\end{equation}
where $ N $ is the batch size, $ \lambda $ is the margin for better similarity separation, and $ [x]_{+} \equiv \max (x, 0) $.
\figurename~\ref{framework}(a) shows the conventional pairwise-based ITR model.
Most current ITR models follow this framework.

\subsubsection{Definition of NDCG} 
This paper uses the ranking metric, Normalised Discounted Cumulative Gain~(NDCG), as the optimization objective of ITR.
Before introducing NDCG, we introduce Discounted Cumulative Gain~(DCG) first.
DCG accumulated at a particular rank position $ p $ is defined as:
\begin{equation}
	\text{DCG}_{p} = 
	\sum_{i=1}^{p}
	\frac{2^{rel_{i}} - 1}
	{\log_{2}(1 + i)},
\end{equation}
where $ rel_{i} $ is the relevance score of the result at position $ i $. 
The premise of DCG is that highly relevant samples ranked low in the retrieval result list should be penalized.
Retrieval result lists vary in length depending on the query.
Comparing the performance from one query to the next cannot be consistently achieved using DCG alone, so the cumulative gain at each position for a chosen value of $ p $ should be normalized across queries.
This is done by ranking all relevant samples in the dataset by their relevance scores, producing the maximum possible DCG through position $ p $, also called the Ideal DCG~(IDCG) through that position.
For a query, the Normalized DCG (NDCG) is computed as:
\begin{equation}
	\text{NDCG}_{p} 
	= \frac{\text{DCG}_{p}}{{\text{IDCG}_{p}}},
\end{equation}
where
\begin{equation}
	\text{IDCG}_{p} =
	\max_{i} \text{DCG}_{p}	=
	\sum_{i=1}^{|REL_{p}|}
	\frac{2^{rel_{i}} - 1}
	{\log_{2}(1 + i)},
\end{equation}
$ REL_{p} $ represents the sample list ordered by their relevance scores in the dataset up to position $ p $.

\subsection{Image-Text Retrieval Framework Integrating Listwise Ranking}
\figurename~\ref{framework} shows our proposed ITR framework integrating listwise ranking.
\figurename~\ref{framework}(a) shows the conventional pairwise-based ITR model.
\figurename~\ref{framework}(b) is our proposed listwise ranking approach, which mainly includes two parts: (a) Relevance Score Calculation~(RSC) module and (b) Smooth-NDCG~(S-NDCG) loss.

Listwise ranking optimizes the entire ranking list based on the relevance score.
Therefore, listwise ranking can rank all candidate samples by relevance, whether they are positive or negative samples.
However, the challenge is that the ITR dataset only annotates positive and negative pairs.
The relevance scores between anchor and negative samples are not annotated.
To address this challenge, we first propose the RSC module. 
The RSC module approximates the relevance score by calculating the text similarity between captions, which enables listwise ranking to be applied in ITR.

For the design of the listwise loss function, we choose NDCG as the optimization objective.
However, NDCG is non-differentiable and cannot be optimized using gradient descent.
To this end, we transform the non-differentiable NDCG into a differentiable listwise loss, named Smooth-NDCG~(S-NDCG).
As shown in \figurename~\ref{framework}, the RSC module and S-NDCG loss constitute our listwise ranking approach, which is independent of the conventional pairwise-based ITR model.
Therefore, listwise ranking can be plug-and-play integrated into pairwise-based ITR models to obtain more user-friendly retrieval results.

\begin{figure}[t]
	\centering
	\includegraphics[width=\linewidth]{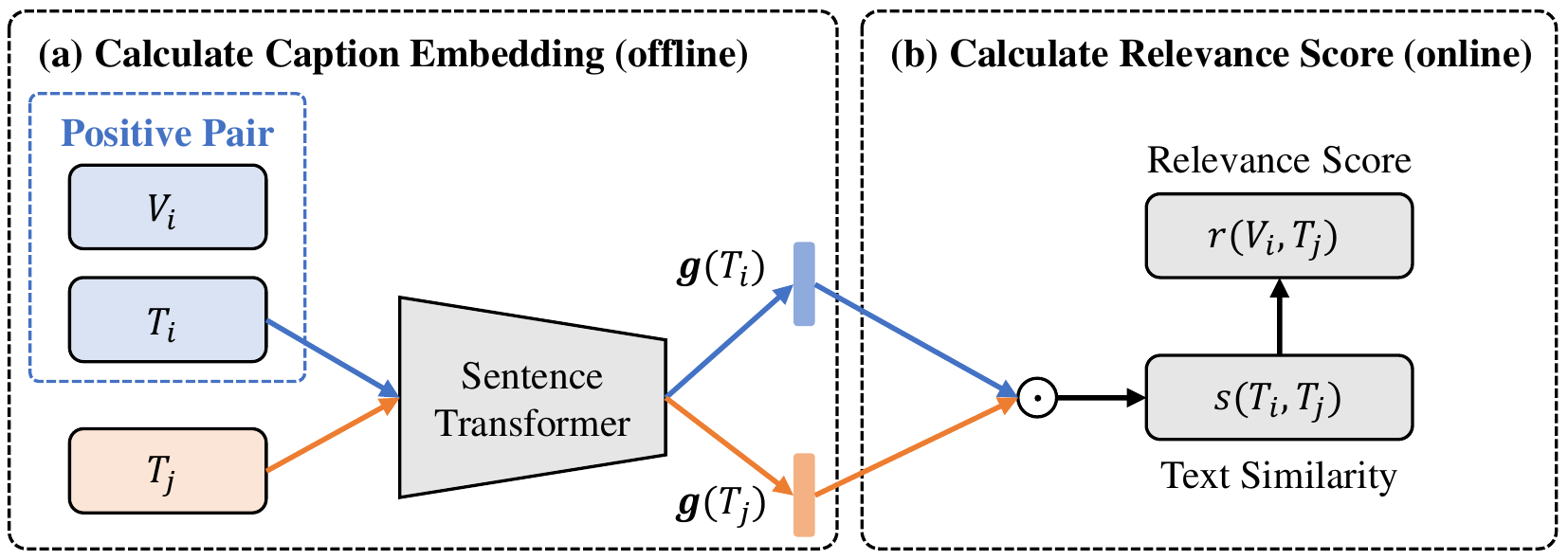}
	\caption{Relevance Score Calculation~(RSC) module.}
	\label{RSC}
\end{figure}
\subsection{Relevance Score Calculation}
It is necessary to obtain the relevance score between the query and the candidate before applying the listwise ranking.
To this end, we propose the Relevance Score Calculation~(RSC) module, as shown in \figurename~\ref{RSC}.
For the ITR task, the query and candidate modalities are different.
Relevance scores across modalities are not easy to measure.
But with the development of the field of Natural Language Processing~(NLP), the measurement of text similarity has approached human levels.
Therefore, we exploit the similarity of captions to approximate the relevance score.
Specifically, to calculate the relevance score $ r(V_{i}, T_{j}) $ between the image $ V_{i} $ and the caption $ T_{j} $, we first need to find the positive caption $ T_{i} $ related to the image $ V_{i} $.
Then we compute the text similarity $ s(T_{i}, T_{j}) $ between $ T_{i} $ and $ T_{j} $.
Finally, we use $ s(T_{i}, T_{j}) $ to approximate the relevance score $ r(V_{i}, T_{j}) $.

As shown in \figurename~\ref{RSC}, our RSC module is divided into two parts: (a)~calculating caption embeddings and (b)~calculating relevance scores.
We adopt the Sentence-Transformers framework~\cite{reimers2019sentence} to calculate caption embeddings.
Sentence-Transformers is a Python framework for state-of-the-art text embeddings.
Text similarity between these embeddings can be computed using a simple similarity metric, \textit{e.g.}, cosine similarity.
Specifically, to calculate the relevance score $ r(V_{i}, T_{j}) $ between the image $ V_{i} $ and the caption $ T_{j} $, we first find the positive caption $ T_{i} $ related to the image $ V_{i} $.
We utilize Sentence-Transformers $ \boldsymbol{g}(\cdot) $ to compute the embeddings of $ T_{i} $ and $ T_{j} $, denoted as $ \boldsymbol{g}(T_{i}) $ and $ \boldsymbol{g}(T_{j}) $, respectively.
Then we use the cosine similarity to calculate the text similarity between $ \boldsymbol{g}(T_{i}) $ and $ \boldsymbol{g}(T_{j}) $:
\begin{equation} \label{relevance_score}
	s(T_{i}, T_{j})
	= \dfrac
	{
		\boldsymbol{g}(T_{i})^{\top} 
		\boldsymbol{g}(T_{j})
	}
	{
		\left \Vert
		\boldsymbol{g}(T_{i})
		\right \Vert 
		\left \Vert
		\boldsymbol{g}(T_{j})
		\right \Vert
	}.
\end{equation}
$ s(T_{i}, T_{j}) \in [-1, 1] $.
We normalize $ s(T_{i}, T_{j}) $ to $ [0, 1] $ to obtain the approximate relevance score:
\begin{equation}
	r(V_{i}, T_{j})
	= \dfrac{1 + s(T_{i}, T_{j})}{2}.
\end{equation}
Note that caption embeddings can be precomputed offline. So the calculation process of caption embedding does not increase the complexity of ITR model training.
Calculating the relevance score during online training only requires calculating a simple cosine similarity.
The RSC module adds negligible computational overhead to the training process.

\begin{table}[t]
	\caption{Comparisons of relevance scores computed by our RSC module and the original pairwise annotations.}
	\setlength\tabcolsep{7pt}
	\begin{center}
		\begin{tabular}{ccccccccccccc}
			\toprule[1pt]
			\multirow{2}*{Relevance Score}
			& \multicolumn{2}{c}{Image-to-Text} & & \multicolumn{2}{c}{Text-to-Image} \\
			\cline{2-3}\cline{5-6}
			\specialrule{0em}{2pt}{0pt}
			~ & mAP@R & R-P & & mAP@R & R-P \\
			\hline
			
			\specialrule{0em}{2pt}{0pt}
			Pairwise Annotation & 31.33 & 31.41 & & 13.62 & 13.67 \\
			\rowcolor{black!10}
			\textbf{RSC} & \textbf{39.93} & \textbf{43.73} & & \textbf{37.53} & \textbf{44.94} \\
			
			\bottomrule[1pt]
		\end{tabular}
	\end{center}
	\label{relevance}
\end{table}
The relevance scores are approximated by our RSC module and may not be exact.
There is a question: 
\textit{how accurate is the relevance score, and can it guide the ITR model optimization?}
To this end, we conduct an experiment to prove that the relevance score can guide model optimization.
We choose the ECCV Caption dataset~\cite{chun2022eccv} for the experiment.
ECCV Caption is a machine-and-human-verified test set of the MS-COCO dataset.
In addition to pairwise positive pairs, the ECCV Caption dataset also annotates false negative pairs.
These false negative pairs are semantically related but mislabeled as negative.
Compared with the pairwise dataset, the ECCV Caption dataset provides more relevance annotations.
The experimental results on the ECCV Caption dataset 
We perform image-to-text and text-to-image ranking based on relevance scores.
We compare the relevance scores computed by the RSC module with the original pairwise annotations.
For pairwise annotations, we set the relevance score to 1 for positive pairs and 0 for negative pairs.
MAP@R and R-Precision~(R-P) are used to evaluate the ranking performance of the relevance scores.
As shown in \tablename~\ref{relevance}, the ranking performance of the relevance score computed by the RSC module is much higher than that of pairwise annotations.
The original pairwise annotations of the dataset can only reflect that the query is related to the positive samples.
Regardless of their relevance to the query, all negative samples are indiscriminately considered irrelevant.
Therefore, pairwise annotation misses a large number of correlations.
The relevance scores computed by our RSC module complement these missing correlations.
The experimental results in \tablename~\ref{relevance} show that the relevance score calculated by our RSC module supplements a large number of correct correlations, which can guide the optimization of the ITR model.
can reflect the quality of relevance ranking.

\subsection{Smooth-NDCG}
Since NDCG is a statistic about ranking, we first need to transform NDCG into a function of the image-text similarities.
To simplify the representation, we abbreviate relevance score $ r(V_{i}, T_{j}) $ as $ r_{i,j} $ in the following. 
In ITR, the DCG of query image $ V_{i} $ in a batch is defined as:
\begin{equation}
	\text{DCG}(V_{i}) =
	\sum_{j=1}^{N}
	\frac{ 2^{r_{i,j}} - 1 }
	{ \log_{2}(1 + \pi_{i,j}) },
\end{equation}
where $ N $ is the batch size, $ r_{i,j} $ is the relevance score between $ V_{i} $ and $ T_{j} $, $ \pi_{i,j} $ is the position of $ T_{j} $ in the ranked list of query $ V_{i} $. 
We use an ranking indicator function $ \mathbb{I} \{ x \} $ to transform $ \pi_{i,j} $ into a function of similarities:
\begin{equation} \label{pi}
	\pi_{i,j} = 1 + 
	\sum_{k = 1, k \neq j}^{N}
	\mathbb{I} \{ s_{i,j} - s_{i,k} < 0 \},
\end{equation}
where
\begin{equation}
	\mathbb{I} \{ x \} =
	\begin{cases}
		1 ,& \text{if $ x $ is true}, \\
		0 ,& \text{otherwise}.
	\end{cases}
\end{equation}
Similarly, we can define IDCG as:
\begin{equation}
	{\rm IDCG}(V_{i}) =
	\sum_{j=1}^{N}
	\frac{2^{r_{i,j}} - 1}
	{\log_{2}(1 + \pi^{*}_{i,j})},
\end{equation}
where $ \pi^{*}_{i,j} $ is the position of $ T_{j} $ in the ideal ranked list of query $ V_{i} $, which is a function of the relevance scores:
\begin{equation}
	\pi^{*}_{i,j} = 1 + 
	\sum_{k = 1, k \neq j}^{N}
	\mathbb{I} \{ r_{i,j} - r_{i,k} < 0 \}.
\end{equation}

\begin{figure}[t]
	\centering
	\subfigure[$ H(x) $]{
		\label{step}
		\includegraphics[width=0.465\linewidth]{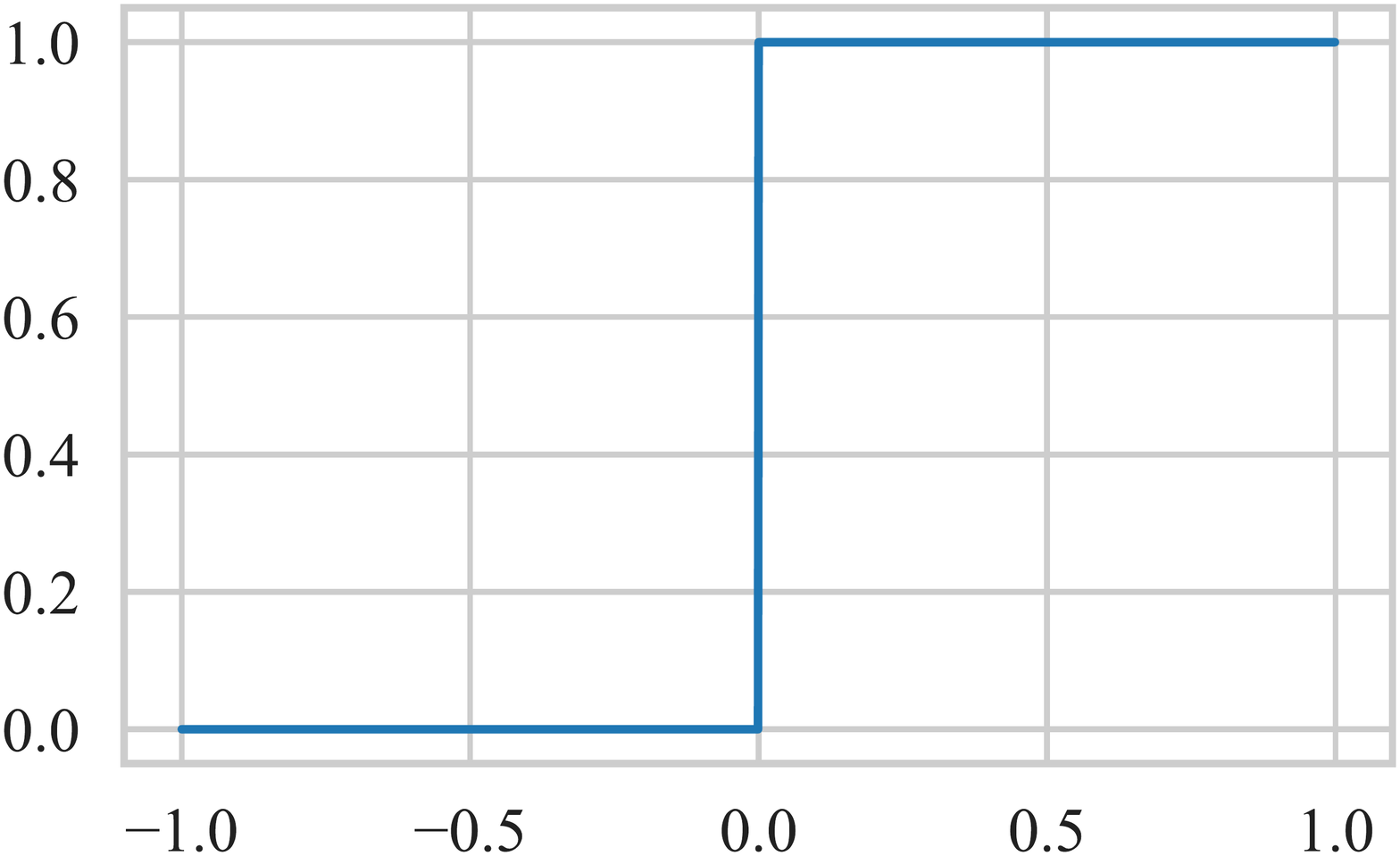}
	}
	\subfigure[$ \delta(x) = \dfrac{ \mathrm{d} H(x) }{ \mathrm{d} x } $]{
		\label{step_derivative}
		\includegraphics[width=0.465\linewidth]{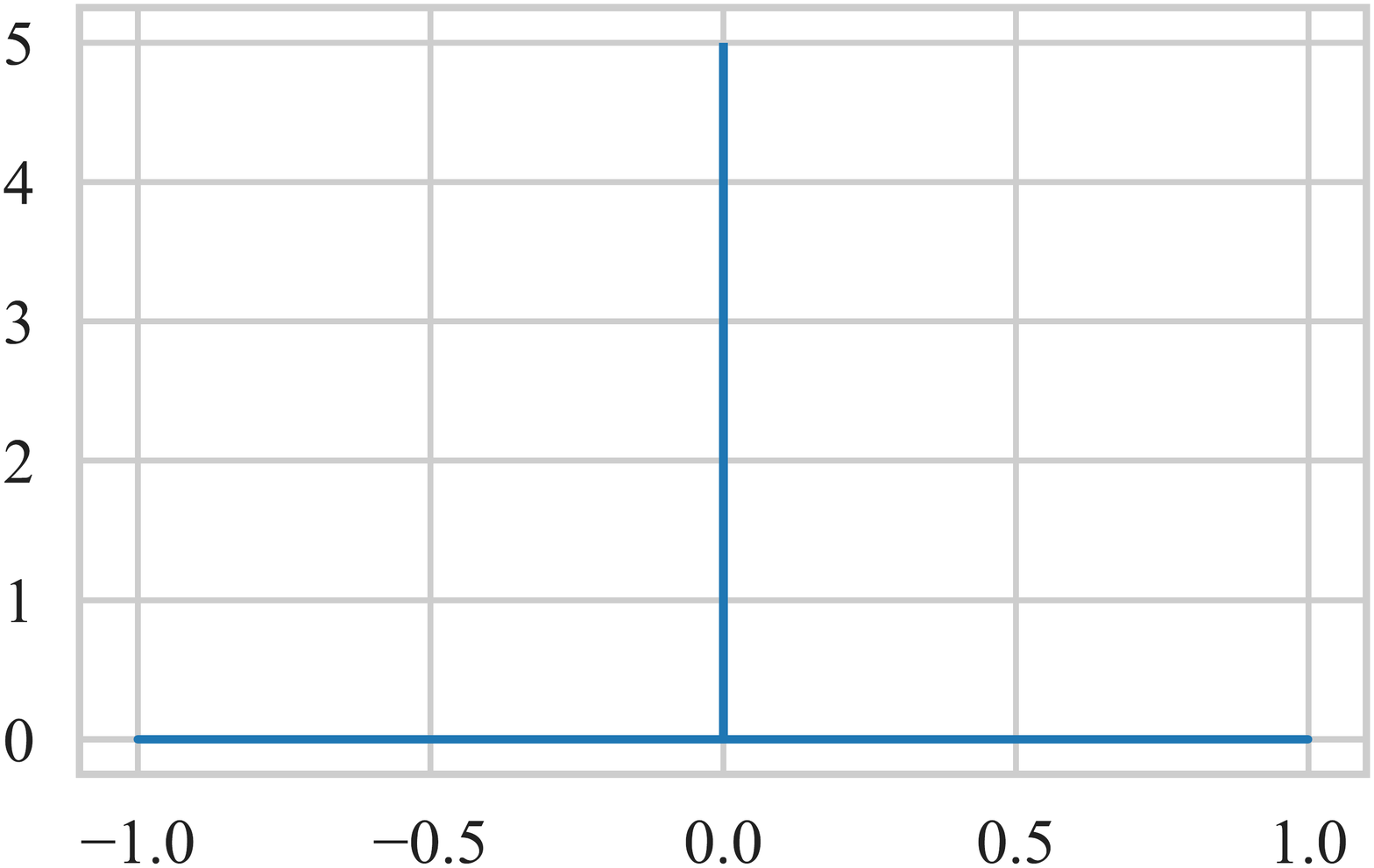}
	}
	\subfigure[$ \sigma(x; \tau) $]{
		\label{sigmoid}
		\includegraphics[width=0.465\linewidth]{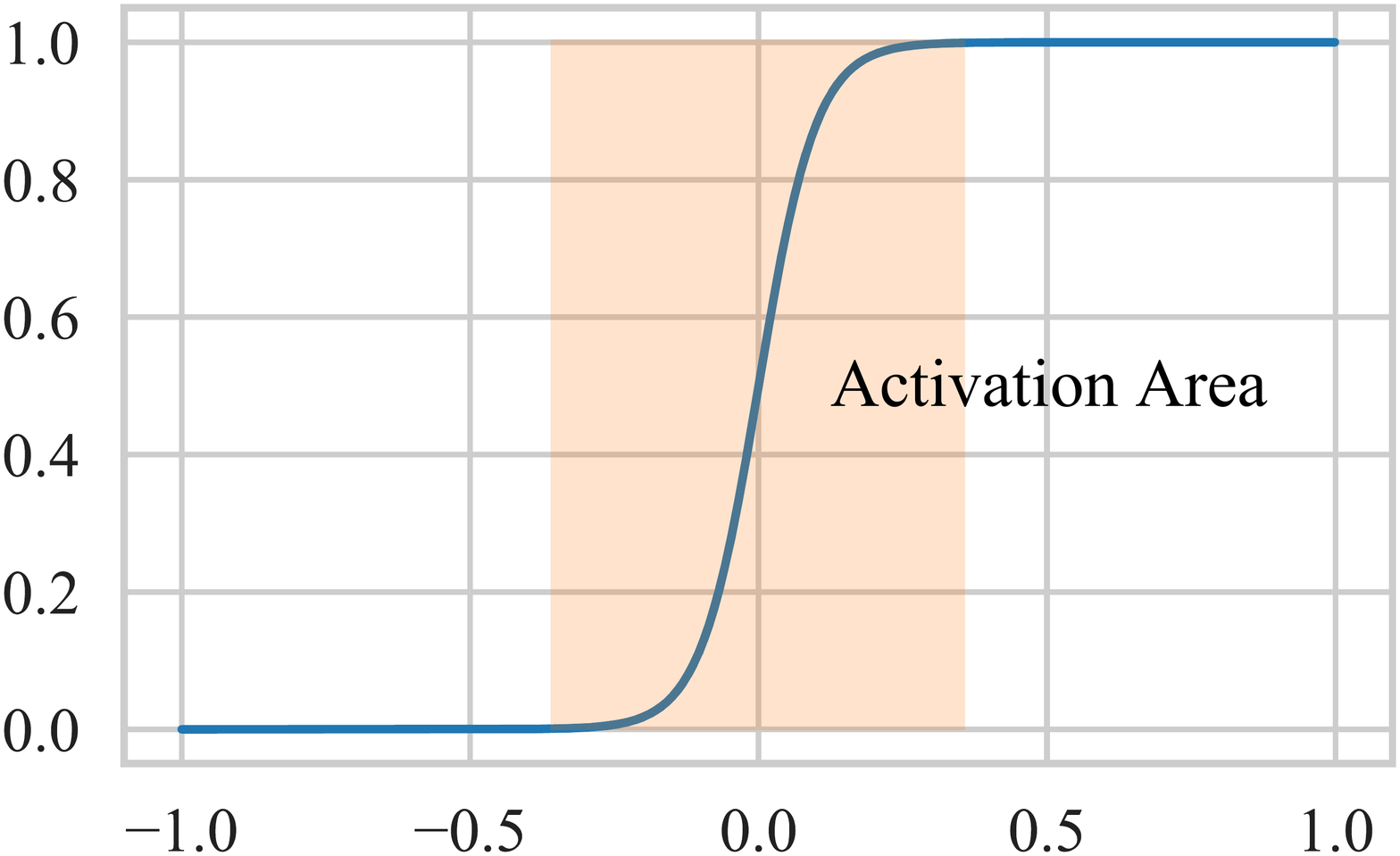}
	}
	\subfigure[$ \dfrac{ \mathrm{d} \sigma(x; \tau) }{ \mathrm{d} x } $]{
		\label{sigmoid_derivative}
		\includegraphics[width=0.465\linewidth]{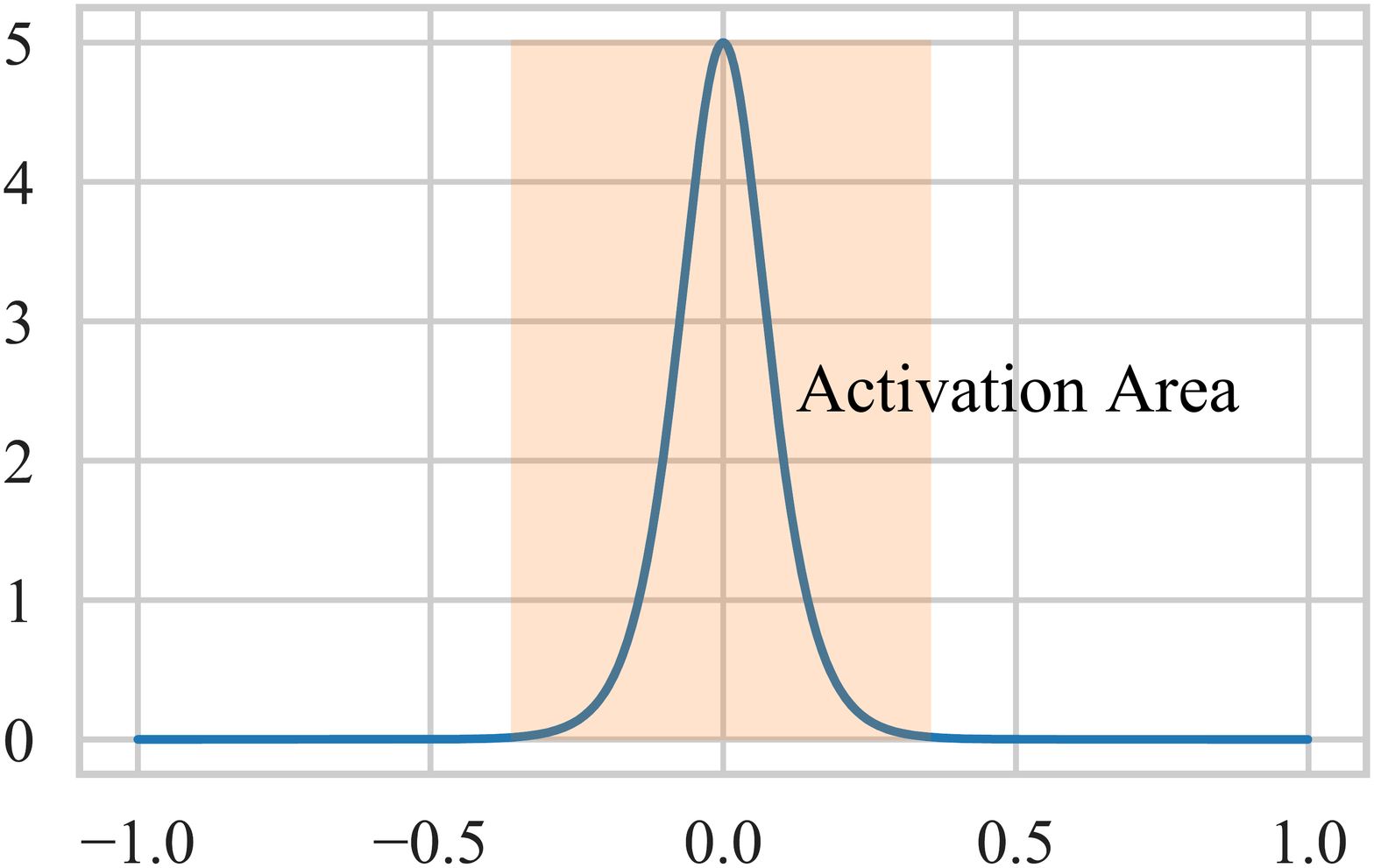}
	}
	\caption{Ranking indicator function and its smooth approximation functions.}
	\label{smooth}
\end{figure}
The particular indicator function $ \mathbb{I} \{ x \} $ is a Heaviside step function $ H(x) $, as shown in \figurename~\ref{step}.
The Dirac delta function $ \delta(x) $ is the derivative of $ H(x) $:
\begin{equation}
	\delta(x) 
	= \dfrac{ \mathrm{d} H(x) }{ \mathrm{d} x },
\end{equation}
as shown in \figurename~\ref{step_derivative}.
Unfortunately, $ H(x) $ is non-continuous and non-differentiable.
Due to this, NDCG cannot be optimized directly with gradient descent.
Brown~\textit{et al.} \cite{brown2020smooth} proposed to use sigmoid function $ \sigma(x; \tau) $ to replace $ H(x) $:
\begin{equation}
	\sigma(x; \tau) 
	= \frac{1}{1 + e^{\frac{-x}{\tau}}},
\end{equation}
where the $ \tau $ refers to the temperature coefficient adjusting the sharpness.
The smaller $ \tau $ is, the closer $ \sigma(x; \tau) $ is to $ H(x) $.
$ \sigma(x; \tau) $ and its derivative are shown in \figurename~\ref{sigmoid} and \figurename~\ref{sigmoid_derivative}.
$ \sigma(x; \tau) $ is a continuously differentiable function.
The orange area in \figurename~\ref{sigmoid_derivative} where the derivative is larger than $ 0 $ is called the activation area, which can provide gradients for ranking optimization.
The larger $ \tau $, the larger the activation area, and the larger the approximation error. 
Therefore, there is a trade-off between the size of the activation area and the approximation error.

Substituting $ \sigma(x; \tau) $ into Eq.~\eqref{pi}, the position function can be approximated as:
\begin{equation} \label{position}
	\widehat{\pi}_{i,j} = 1 + 
	\sum_{k = 1, k \neq j}^{N} 
	\sigma ( s_{i,j} - s_{i,k} ).
\end{equation}
The approximated DCG can be expressed as:
\begin{equation}
	\widehat{\text{DCG}}(V_{i}) =
	\sum_{j=1}^{N}
	\frac{2^{r_{i,j}} - 1}
	{\log_{2}(1 + \widehat{\pi}_{i,j})}.
\end{equation}
We can get the approximated NDCG:
\begin{equation}
	\widehat{\text{NDCG}}(V_{i}) = \frac{\widehat{\text{DCG}}(V_{i})}{{\rm IDCG}(V_{i})}.
\end{equation}
Since ITR includes two types of retrieval from image-to-text~($ v \rightarrow t $) and text-to-image~($ t \rightarrow v $), the listwise loss function Smooth-NDCG~(S-NDCG) is denoted as:
\begin{equation}
	\mathcal{L}_{\text{S-NDCG}} 
	= \mathcal{L}_{\text{S-NDCG}}^{v \rightarrow t}
	+ \mathcal{L}_{\text{S-NDCG}}^{t \rightarrow v}.
\end{equation}
$ \mathcal{L}_{\text{S-NDCG}}^{v \rightarrow t} $ takes the form of:
\begin{equation}
	\mathcal{L}_{\text{S-NDCG}}^{v \rightarrow t} 
	= \dfrac{1}{N}
	\sum_{i=1}^{N} 
	\left(
	1 - \widehat{\text{NDCG}}(V_{i})
	\right).
\end{equation}
$ \mathcal{L}_{\text{S-NDCG}}^{t \rightarrow v} $ takes the form of:
\begin{equation}
	\mathcal{L}_{\text{S-NDCG}}^{t \rightarrow v} 
	= \dfrac{1}{N}
	\sum_{i=1}^{N} 
	\left(
	1 - \widehat{\text{NDCG}}(T_{i})
	\right).
\end{equation}

For end-to-end training, NDCG cannot be globally optimized.
Training a deep learning model requires dividing the data into batches.
There is a gap between the NDCG on each batch and the NDCG on the entire dataset.
Optimizing S-NDCG within a batch does not generalize well to the entire dataset.
Pairwise losses generally have better generalization.
Triplet loss introduces a margin for similarity separation to enhance generalization.
In order to make up for the shortcoming of S-NDCG in generalization, we jointly use triplet loss and S-NDCG as the optimization objective:
\begin{equation}
	\mathcal{L}
	= \mathcal{L}_{\text{Triplet}}
	+ \mathcal{L}_{\text{S-NDCG}}.
\end{equation}
Triplet loss can compensate for the generalization of S-NDCG, but it cannot rank all samples by relevance.
Integrating listwise ranking into the pairwise-based ITR model can not only guarantee generalization, but also achieve the relevance ranking of all candidates.

\subsection{Computational Overhead} \label{overhead}
Our listwise ranking can be integrated into conventional pairwise-based ITR models without significantly increasing computational overhead.
To this end, we conduct in-depth analysis in two areas: training and testing.
We calculate the additional computational overhead of listwise ranking.

\subsubsection{Training}
Compared with the pairwise-based ITR model, the integrated listwise ranking increases computational overhead in two aspects: relevance score calculation and S-NDCG loss function calculation.

The RSC module calculates the relevance score in real time during training.
As shown in \figurename~\ref{RSC}, the RSC module is divided into two parts: (a)~calculating caption embeddings and (b)~calculating relevance scores.
Caption embeddings can be precomputed offline. 
Therefore, calculating caption embeddings does not increase the computational overhead of training.
Eq.~\eqref{relevance_score} shows that calculating the relevance score during online training only requires calculating a simple cosine similarity.
When the batch size is $ N $, there are $ N^{2} $ relevance scores that need to be calculated.
Computing all relevance scores requires $ N^{2} $ dot product operations.
Therefore, calculating the relevance score during training adds $ N^{2} $ additional dot product operations.
For deep neural network training, $ N^{2} $ dot product operations are insignificant.
The RSC module adds negligible computational overhead to the training process.

As shown in Eq.~\eqref{triplet}, computing the triplet loss requires computing $ (s_{i,j} - s_{i,i}) $, which is the difference between negative and positive pair similarities. 
When the batch size is $ N $, there are $ N $ anchors in a batch.
Each anchor has a corresponding positive sample and $ N-1 $ negative samples.
The complexity of calculating the triplet loss is $ O(N^{2}) $.
As shown in Eq.~\eqref{position}, computing S-NDCG loss requires computing $ (s_{i,j} - s_{i,k}) $, which is the difference between any two image-text pairs.
Each anchor corresponds to $ N $ image-text pairs.
There are $ N^{2} $ similarity differences between any two image-text pairs.
The complexity of calculating S-NDCG loss is $ O(N^{3}) $.
The computational complexity of S-NDCG is an order of magnitude higher than triplet loss.
But we can make up for the disadvantage of S-NDCG in complexity through engineering implementation.
The calculation of $ (s_{i,j} - s_{i,k}) $ can be implemented as the difference between two similarity matrices:
\begin{equation}
	\begin{aligned}
		\Delta \mathbf{S}_{i}
		& = \begin{bmatrix} 
			s_{i,1} - s_{i,1}  &  \dots & s_{i,N} - s_{i,1} \\
			\vdots  &  \ddots  & \vdots \\ 
			s_{i,1} - s_{i,N}  &  \dots & s_{i,N} - s_{i,N} 
		\end{bmatrix} \\
		& = \begin{bmatrix} 
			s_{i,1}  &  \dots   & s_{i,N} \\
			\vdots   &  \ddots  & \vdots \\ 
			s_{i,1}  &  \dots   & s_{i,N} 
		\end{bmatrix}
		- 
		\begin{bmatrix} 
			s_{i,1} &  \dots   &   s_{i,1} \\
			\vdots  &  \ddots  &   \vdots \\
			s_{i,N} &  \dots   &   s_{i,N}
		\end{bmatrix}.
	\end{aligned}
\end{equation}
Matrix operations are suitable for GPU parallel acceleration operations.
In addition, the trainable parameters of the ITR model are mainly the backbone network parameters of the image and text encoders.
Both our RSC module and S-NDCG loss function are independent of the backbone network, so it does not add too much computational overhead.
Our subsequent experiments verify that integrating our listwise ranking only adds a small amount of training time.

\subsubsection{Testing}
Our listwise ranking is independent of the ITR model architecture.
In the testing phase, the ITR model only needs to calculate the similarity between images and captions and rank them.
The RSC module and S-NDCG loss do not need to be involved in the testing process.
Therefore, integrating listwise ranking does not increase computational overhead in the testing phase.

We then conduct computational efficiency experiments in the following experiments.
The experimental results are consistent with our analysis.

\section{Experiments}
\subsection{Datasets and Experiment Settings}
\subsubsection{Datasets}
We use three benchmarks for evaluation:
Flickr30K~\cite{young2014image}, MS-COCO~\cite{lin2014microsoft} and ECCV Caption~\cite{chun2022eccv}.
\begin{itemize}[leftmargin=*]
	\item 
	\textbf{Flickr30K} contains 31,000 images; each image is annotated with 5 sentences.
	We use 29,000 images for training, 1,000 images for validation, and 1,000 images for testing.
	\item 
	\textbf{MS-COCO} contains 123,287 images; each image is annotated with 5 sentences.
	We use 113,287 images for training, 5,000 images for validation, and 5,000 images for testing.
	We report results on both 1,000 test images~(averaged over five folds) and the full 5,000 test images.
	\item 
	\textbf{ECCV Caption} is a machine-and-human-verified test set of the MS-COCO dataset.
	ECCV Caption provides $ \times 3.6 $ positive image-to-text associations and $ \times 8.5 $ text-to-image associations compared to the original MS-COCO dataset.
	We use the MS-COCO training set to train the model and use the ECCV Caption dataset for testing.
\end{itemize}

\begin{table}[t] 
	\caption{Experimental results on the Flickr30K dataset}
	\setlength\tabcolsep{2.3pt}
	\begin{center}
		\begin{tabular}{lcccccccccccc}
			\toprule[1pt]
			\multirow{2}*{Model}
			& \multicolumn{3}{c}{Image-to-Text} & & \multicolumn{3}{c}{Text-to-Image} & \multirow{2}*{RSUM} \\
			\cline{2-4}\cline{6-8}
			\specialrule{0em}{2pt}{0pt}
			~ & R@1 & R@5 & R@10 & & R@1 & R@5 & R@10 \\
			\hline
			
			\specialrule{0em}{2pt}{0pt}
			\multicolumn{6}{l}{(Local-Level Alignment)} \\
			SCAN$ _{\text{(\textit{ECCV}'18)}} $~\cite{lee2018stacked} & 67.4 & 90.3 & 95.8 & & 48.6 & 77.7 & 85.2 & 465.0 \\
			CAMP$ _{\text{(\textit{ICCV}'19)}} $~\cite{wang2019camp} & 68.1 & 89.7 & 95.2 & & 51.5 & 77.1 & 85.3 & 466.9 \\
			BFAN$ _{\text{(\textit{MM}'19)}} $~\cite{liu2019focus} & 68.1 & 91.4 & 95.9 & & 50.8 & 78.4 & 85.8 & 470.4 \\
			PFAN$  _{\text{(\textit{IJCAI}'19)}}$~\cite{wang2019position} & 70.0 & 91.8 & 95.0 & & 50.4 & 78.7 & 86.1 & 472.0 \\
			PFAN++$ _{\text{(\textit{TMM}'20)}} $~\cite{wang2020pfan++} & 70.1 & 91.8 & 96.1 & & 52.7 & 79.9 & 87.0 & 477.6 \\
			CAAN$ _{\text{(\textit{CVPR}'20)}} $~\cite{zhang2020context} & 70.1 & 91.6 & 97.2 & & 52.8 & 79.0 & 87.9 & 478.6 \\
			RRTC$ _{\text{(\textit{TCSVT}'21)}} $~\cite{wu2021region} & 72.7 & 93.8 & 96.8 & & 54.2 & 79.4 & 86.1 & 483.0 \\
			MMCA$ _{\text{(\textit{CVPR}'20)}} $~\cite{wei2020multi} & 74.2 & 92.8 & 96.4 & & 54.8 & 81.4 & 87.8 & 487.4 \\
			IMRAM$ _{\text{(\textit{CVPR}'20)}} $~\cite{chen2020imram} &  74.1 & 93.0 & 96.6 & & 53.9 & 79.4 & 87.2 & 484.2 \\
			GSMN$ _{\text{(\textit{CVPR}'20)}} $~\cite{liu2020graph} & 76.4 & 94.3 & 97.3 & & 57.4 & 82.3 & 89.0 & 496.8 \\
			UARDA$ _{\text{(\textit{TMM}'22)}} $~\cite{zhang2022unified} & 77.8 & 95.0 & 97.6 & & 57.8 & 82.9 & 89.2 & 500.3 \\
			DREN$ _{\text{(\textit{TCSVT}'22)}} $~\cite{yang2022dual} & 78.8 & 95.0 & 97.5 & & 58.8 & 83.8 & 89.5 & 503.4 \\
			CMCAN$ _{\text{(\textit{AAAI}'22)}} $~\cite{zhang2022show} & 79.5 & 95.6 & 97.6 & & 60.9 & 84.3 & 89.9 & 507.8 \\
			DRCE$ _{\text{(\textit{TCSVT}'23)}} $~\cite{wang2022dual} & 81.0 & 96.0 & 97.7 & & 60.0 & 84.1 & 89.6 & 508.4 \\
			NAAF$ _{\text{(\textit{CVPR}'22)}} $~\cite{zhang2022negative} & 81.9 & 96.1 & 98.3 & & 61.0 & 85.3 & 90.6 & 513.2 \\
			%			DIME$ _{\text{(\textit{SIGIR}'21)}} $~\cite{qu2021dynamic} & 81.0 & 95.9 & 98.4 & & 63.6 & 88.1 & 93.0 & 520.0 \\
			\hline
			
			\specialrule{0em}{2pt}{0pt}
			SAF$ _{\text{(\textit{AAAI}'21)}} $~\cite{diao2021similarity} & 73.7 & 93.3 & 96.3 & & 56.1 & 81.5 & 88.0 & 488.9 \\
			\rowcolor{black!10}
			\textbf{+ \textit{Listwise}} & 77.4 & 94.1 & 97.2 & & 57.3 & 83.5 & 89.2 & \textbf{498.7} \\
			\hline
			\specialrule{0em}{2pt}{0pt}
			SGR$ _{\text{(\textit{AAAI}'21)}} $~\cite{diao2021similarity} & 75.2 & 93.3 & 96.6 & & 56.2 & 81.0 & 86.5 & 488.8 \\
			\rowcolor{black!10}
			\textbf{+ \textit{Listwise}} & 77.2 & 94.5 & 97.5 & & 57.9 & 83.5 & 88.8 & \textbf{499.4} \\
			\hline
			\specialrule{0em}{2pt}{0pt}
			SGRAF$ _{\text{(\textit{AAAI}'21)}} $~\cite{diao2021similarity} & 77.8 & 94.1 & 97.4 & & 58.5 & 83.0 & 88.8 & 499.6 \\
			\rowcolor{black!10}
			\textbf{+ \textit{Listwise}} & 80.5 & 95.2 & 97.7 & & 60.6 & 85.0 & 90.1 & \textbf{509.1} \\
			\hline
			
			\specialrule{0em}{2pt}{0pt}
			\multicolumn{6}{l}{(Global-Level Alignment)} \\
			CVSE$ _{\text{(\textit{ECCV}'20)}} $~\cite{wang2020consensus} & 73.5 & 92.1 & 95.8 & & 52.9 & 80.4 & 87.8 & 482.5 \\
			VSRN$ _{\text{(\textit{ICCV}'19)}} $~\cite{li2019visual} & 71.3 & 90.6 & 96.0 & & 54.7 & 81.8 & 88.2 & 482.6 \\
			VSRN++$ _{\text{(\textit{TPAMI}'22)}} $~\cite{li2022image} & 79.2 & 94.6 & 97.5 & & 60.6 & 85.6 & 91.4 & 508.9 \\
			DSRAN$ _{\text{(\textit{TCSVT}'21)}} $~\cite{wen2021learning} & 80.5 & 95.5 & 97.9 & & 59.2 & 86.0 & 91.9 & 511.0 \\
			VSE$\infty$ $ _{\text{(\textit{CVPR}'21)}} $~\cite{chen2021learning} & 81.7 & 95.4 & 97.6 & & 61.4 & 85.9 & 91.5 & 513.5 \\
			\hline
			
			\specialrule{0em}{2pt}{0pt}
			VSE++$ _{\text{(\textit{BMVC}'18)}} $~\cite{faghri2018vse++} & 52.9 & 80.5 & 87.2 & & 39.6 & 70.1 & 79.5 & 409.8 \\
			VSE++$ ^{\dag} $ & 70.4 & 90.7 & 95.2 & & 51.6 & 78.6 & 86.2 & 472.7 \\
			\rowcolor{black!10}
			\textbf{+ \textit{Listwise}} & 70.6 & 91.9 & 95.6 & & 53.0 & 79.9 & 86.8 & \textbf{477.7} \\
			\hline
			
			\specialrule{0em}{2pt}{0pt}
			VSE$ \infty $$ _{\text{(\textit{CVPR}'21)}} $~\cite{chen2021learning} & 81.7 & 95.4 & 97.6 & & 61.4 & 85.9 & 91.5 & 513.5 \\
			\rowcolor{black!10}
			\textbf{+ \textit{Listwise}} & 81.0 & 96.0 & 98.2 & & 62.6 & 87.2 & 92.5 & 517.4 \\
			\rowcolor{black!15}
			\textbf{+ \textit{Listwise*}} & 82.6 & 96.4 & 98.5 & & 63.5 & 88.0 & 92.9 & \textbf{521.9} \\
			
			\bottomrule[1pt]
		\end{tabular}
	\end{center}
	\label{f30k}
\end{table}
\begin{table}[t] 
	\caption{Experimental results on the MS-COCO 1K test set}
	\setlength\tabcolsep{2.3pt}
	\begin{center}
		\begin{tabular}{lcccccccccccc}
			\toprule[1pt]
			\multirow{2}*{Model}
			& \multicolumn{3}{c}{Image-to-Text} & & \multicolumn{3}{c}{Text-to-Image} & \multirow{2}*{RSUM} \\
			\cline{2-4}\cline{6-8}
			\specialrule{0em}{2pt}{0pt}
			~ & R@1 & R@5 & R@10 & & R@1 & R@5 & R@10 \\
			\hline
			
			\specialrule{0em}{2pt}{0pt}
			\multicolumn{6}{l}{(Local-Level Alignment)} \\
			CAMP$ _{\text{(\textit{ICCV}'19)}} $~\cite{wang2019camp} & 72.3 & 94.8 & 98.3 & & 58.5 & 87.9 & 95.0 & 506.8 \\ 
			SCAN$ _{\text{(\textit{ECCV}'18)}} $~\cite{lee2018stacked} & 72.7 & 94.8 & 98.4 & & 58.8 & 88.4 & 94.8 & 507.9 \\
			BFAN$ _{\text{(\textit{MM}'19)}} $~\cite{liu2019focus} & 74.9 & 95.2 & 98.3 & & 59.4 & 88.4 & 94.5 & 510.7 \\
			MMCA$ _{\text{(\textit{CVPR}'20)}} $~\cite{wei2020multi} & 74.8 & 95.6 & 97.7 & & 61.6 & 89.8 & 95.2 & 514.7 \\
			CAAN$ _{\text{(\textit{CVPR}'20)}} $~\cite{zhang2020context} & 75.5 & 95.4 & 98.5 & & 61.3 & 89.7 & 95.2 & 515.6 \\
			IMRAM$ _{\text{(\textit{CVPR}'20)}} $~\cite{chen2020imram} & 76.7 & 95.6 & 98.5 & & 61.7 & 89.1 & 95.0 & 516.6 \\
			RRTC$ _{\text{(\textit{TCSVT}'21)}} $~\cite{wu2021region} & 76.2 & 96.3 & 98.9 & & 61.6 & 89.3 & 94.6 & 516.9 \\
			PFAN$ _{\text{(\textit{IJCAI}'19)}} $~\cite{wang2019position} & 76.5 & 96.3 & 99.0 & & 61.6 & 89.6 & 95.2 & 518.2 \\
			PFAN++$ _{\text{(\textit{TMM}'20)}} $~\cite{wang2020pfan++} & 77.1 & 96.5 & 98.3 & & 62.5 & 89.9 & 95.4 & 519.7 \\
			GSMN$ _{\text{(\textit{CVPR}'20)}} $~\cite{liu2020graph} & 78.4 & 96.4 & 98.6 & & 63.3 & 90.1 & 95.7 & 522.5 \\
			UARDA$ _{\text{(\textit{TMM}'22)}} $~\cite{zhang2022unified} & 78.6 & 96.5 & 98.9 & & 63.9 & 90.7 & 96.2 & 524.8 \\
			%			DIME$ _{\text{(\textit{SIGIR}'21)}} $~\cite{qu2021dynamic} & 78.8 & 96.3 & 98.7 & & 64.8 & 91.5 & 96.5 & 526.6 \\
			NAAF$ _{\text{(\textit{CVPR}'22)}} $~\cite{zhang2022negative} & 80.5 & 96.5 & 98.8 & & 64.1 & 90.7 & 96.5 & 527.2 \\
			CMCAN$ _{\text{(\textit{AAAI}'22)}} $~\cite{zhang2022show} & 81.2 & 96.8 & 98.7 & & 65.4 & 91.0 & 96.2 & 529.3 \\
			\hline
			
			\specialrule{0em}{2pt}{0pt}
			SAF$ _{\text{(\textit{AAAI}'21)}} $~\cite{diao2021similarity} & 76.1 & 95.4 & 98.3 & & 61.8 & 89.4 & 95.3 & 516.3 \\
			\rowcolor{black!10}
			\textbf{+ \textit{Listwise}} & 77.5 & 95.6 & 98.4 & & 62.8 & 89.7 & 95.4 & \textbf{519.4} \\
			\hline
			\specialrule{0em}{2pt}{0pt}
			SGR$ _{\text{(\textit{AAAI}'21)}} $~\cite{diao2021similarity} & 78.0 & 95.8 & 98.2 & & 61.4 & 89.3 & 95.4 & 518.1 \\
			\rowcolor{black!10}
			\textbf{+ \textit{Listwise}} & 78.1 & 96.1 & 98.6 & & 62.7 & 89.8 & 95.4 & \textbf{520.7} \\
			\hline
			\specialrule{0em}{2pt}{0pt}
			SGRAF$ _{\text{(\textit{AAAI}'21)}} $~\cite{diao2021similarity} & 79.6 & 96.2 & 98.5 & & 63.2 & 90.7 & 96.1 & 524.3 \\
			\rowcolor{black!10}
			\textbf{+ \textit{Listwise}} & 80.1 & 96.5 & 98.9 & & 64.3 & 90.8 & 96.1 & \textbf{526.7} \\
			\hline
			
			\specialrule{0em}{2pt}{0pt}
			\multicolumn{6}{l}{(Global-Level Alignment)} \\
			CVSE$ _{\text{(\textit{ECCV}'20)}} $~\cite{wang2020consensus} & 74.8 & 95.1 & 98.3 & & 59.9 & 89.4 & 95.2 & 512.7 \\
			VSRN$ _{\text{(\textit{ICCV}'19)}} $~\cite{li2019visual} & 76.2 & 94.8 & 98.2 & & 62.8 & 89.7 & 95.1 & 516.8 \\
			VSRN++$ _{\text{(\textit{TPAMI}'22)}} $~\cite{li2022image} & 77.9 & 96.0 & 98.5 & & 64.1 & 91.0 & 96.1 & 523.6 \\
			DSRAN$ _{\text{(\textit{TCSVT}'21)}} $~\cite{wen2021learning} & 80.6 & 96.7 & 98.7 & & 64.5 & 90.8 & 95.8 & 527.1 \\
			VSE$\infty$ $ _{\text{(\textit{CVPR}'21)}} $~\cite{chen2021learning} & 79.7 & 96.4 & 98.9 & & 64.8 & 91.4 & 96.3 & 527.5 \\
			\hline
			
			\specialrule{0em}{2pt}{0pt}
			VSE++$ _{\text{(\textit{BMVC}'18)}} $~\cite{faghri2018vse++} & 64.6 & 90.0 & 95.7 & & 52.0 & 84.3 & 92.0 & 478.6 \\
			VSE++$ ^{\dag} $ & 72.8 & 94.6 & 98.1 & & 57.9 & 88.0 & 94.4 & 505.8 \\
			\rowcolor{black!10}
			\textbf{+ \textit{Listwise}} & 73.4 & 94.4 & 98.1 & & 58.8 & 88.4 & 94.6 & \textbf{507.7} \\
			\hline
			
			\specialrule{0em}{2pt}{0pt}
			VSE$ \infty $$ _{\text{(\textit{CVPR}'21)}} $~\cite{chen2021learning} & 79.7 & 96.4 & 98.9 & & 64.8 & 91.4 & 96.3 & 527.5 \\
			\rowcolor{black!10}
			\textbf{+ \textit{Listwise}} & 80.2 & 96.6 & 98.9 & & 65.3 & 91.3 & 96.3 & 528.7 \\
			\rowcolor{black!15}
			\textbf{+ \textit{Listwise*}} & 80.8 & 96.7 & 98.9 & & 66.0 & 91.9 & 96.5 & \textbf{530.8} \\
			
			\bottomrule[1pt]
		\end{tabular}
	\end{center}
	\label{coco1k}
\end{table}
\begin{table}[t] 
	\caption{Experimental results on the MS-COCO 5K test set}
	\setlength\tabcolsep{2.3pt}
	\begin{center}
		\begin{tabular}{lcccccccccccc}
			\toprule[1pt]
			\multirow{2}*{Model}
			& \multicolumn{3}{c}{Image-to-Text} & & \multicolumn{3}{c}{Text-to-Image} & \multirow{2}*{RSUM} \\
			\cline{2-4}\cline{6-8}
			\specialrule{0em}{2pt}{0pt}
			~ & R@1 & R@5 & R@10 & & R@1 & R@5 & R@10 \\
			\hline
			
			\specialrule{0em}{2pt}{0pt}
			\multicolumn{6}{l}{(Local-Level Alignment)} \\
			SCAN$_{\text{(\textit{ECCV}'18)}}$~\cite{lee2018stacked} & 50.4 & 82.2 & 90.0 & & 38.6 & 69.3 & 80.4 & 410.9 \\
			BFAN$ _{\text{(\textit{MM}'19)}} $~\cite{liu2019focus} & 52.9 & 82.8 & 90.6 & & 38.3 & 67.8 & 79.3 & 411.7 \\
			IMRAM$_{\text{(\textit{CVPR}'20)}}$~\cite{chen2020imram} & 53.7 & 83.2 & 91.0 & & 39.7 & 69.1 & 79.8 & 416.5 \\
			DRCE$ _{\text{(\textit{TCSVT}'23)}} $~\cite{wang2022dual} & 56.2 & 83.0 & 90.9 & & 40.3 & 69.5 & 80.6 & 420.5 \\
			NAAF$_{\text{(\textit{CVPR}'22)}}$~\cite{zhang2022negative} & 58.9 & 85.2 & 92.0 & & 42.5 & 70.9 & 81.4 & 430.9 \\
%			DIME$ _{\text{(\textit{SIGIR}'21)}} $~\cite{qu2021dynamic} & 59.3 & 85.4 & 91.9 & & 43.1 & 73.0 & 83.1 & 435.8 \\
%			CMCAN$_{\text{(\textit{AAAI}'22)}}$~\cite{zhang2022show} & 61.5 & - & 92.9 & & 44.0 & - & 82.6 & - \\
			\hline
			
			\specialrule{0em}{2pt}{0pt}
			SAF$ _{\text{(\textit{AAAI}'21)}} $~\cite{diao2021similarity} & 53.3 & - & 90.1 & & 39.8 & - & 80.2 & - \\
			\rowcolor{black!10}
			\textbf{+ \textit{Listwise}} & 54.6 & 83.3 & 91.3 & & 41.1 & 70.3 & 80.7 & \textbf{421.3} \\
			\hline
			\specialrule{0em}{2pt}{0pt}
			SGR$ _{\text{(\textit{AAAI}'21)}} $~\cite{diao2021similarity} & 56.9 & - & 90.5 & & 40.2 & - & 79.8 & - \\
			\rowcolor{black!10}
			\textbf{+ \textit{Listwise}} & 56.9 & 84.0 & 91.1 & & 41.1 & 70.2 & 80.5 & \textbf{423.8} \\
			\hline
			\specialrule{0em}{2pt}{0pt}
			SGRAF$_{\text{(\textit{AAAI}'21)}}$~\cite{diao2021similarity} & 57.8 & - & 91.6 & & 41.9 & - & 81.3 & - \\
			\rowcolor{black!10}
			\textbf{+ \textit{Listwise}} & 59.2 & 85.1 & 92.1 & & 43.0 & 71.7 & 82.0 & \textbf{433.1} \\
			\hline
			
			\specialrule{0em}{2pt}{0pt}
			\multicolumn{6}{l}{(Global-Level Alignment)} \\
			VSRN$_{\text{(\textit{ICCV}'19)}}$~\cite{li2019visual} & 53.0 & 81.1 & 89.4 & & 40.5 & 70.6 & 81.1 & 415.7 \\
			VSRN++$_{\text{(\textit{TPAMI}'22)}}$~\cite{li2022image} & 54.7 & 82.9 & 90.9 & & 42.0 & 72.2 & 82.7 & 425.4 \\
			DSRAN$ _{\text{(\textit{TCSVT}'21)}} $~\cite{wen2021learning} & 57.9 & 85.3 & 92.0 & & 41.7 & 72.7 & 82.8 & 432.4 \\
			VSE$\infty$ $ _{\text{(\textit{CVPR}'21)}} $~\cite{chen2021learning} & 58.3 & 85.3 & 92.3 & & 42.4 & 72.7 & 83.2 & 434.3 \\
			\hline
			
			\specialrule{0em}{2pt}{0pt}
			VSE++$ _{\text{(\textit{BMVC}'18)}} $~\cite{faghri2018vse++} & 41.3 & 71.1 & 81.2 & & 30.3 & 59.4 & 72.4 & 355.7 \\
			VSE++$ ^{\dag} $ & 49.9 & 78.4 & 88.0 & & 35.6 & 65.9 & 77.5 & 395.4 \\
			\rowcolor{black!10}
			\textbf{+ \textit{Listwise}} & 50.5 & 79.5 & 88.1 & & 36.0 & 66.4 & 78.2 & \textbf{398.6} \\
			\hline
			
			\specialrule{0em}{2pt}{0pt}
			VSE$ \infty $$ _{\text{(\textit{CVPR}'21)}} $~\cite{chen2021learning} & 58.3 & 85.3 & 92.3 & & 42.4 & 72.7 & 83.2 & 434.3 \\
			\rowcolor{black!10}
			\textbf{+ \textit{Listwise}} & 59.1 & 85.4 & 92.6 & & 42.9 & 73.4 & 83.4 & 436.9 \\
			\rowcolor{black!15}
			\textbf{+ \textit{Listwise*}} & 60.6 & 86.3 & 92.5 & & 43.9 & 73.9 & 84.0 & \textbf{441.2} \\
			
			\bottomrule[1pt]
		\end{tabular}
	\end{center}
	\label{coco5k}
\end{table}
\subsubsection{Baseline Models}
To demonstrate that listwise ranking can improve the performance of current conventional pairwise-based ITR models, we conduct experiments on the VSE++~\cite{faghri2018vse++}, VSE$ \infty $~\cite{chen2021learning}, and SGRAF~\cite{diao2021similarity} models.
\begin{itemize}[leftmargin=*]
	\item 
	\textbf{VSE++} is the most representative global-level alignment model in ITR.
	A number of subsequent proposed ITR models are variants of VSE++.
	We re-implement VSE++ using Bottom-Up and Top-Down~(BUTD) attention region features~\cite{anderson2018bottom} as our baseline, denoted as VSE++$ ^{\dagger} $. 
	Image features are extracted by a pre-trained Faster R-CNN~\cite{anderson2018bottom} with a ResNet-101~\cite{he2016deep} backbone. 
	Then we map the image features to a set of 1024-dimensional vectors through a fully connected layer.
	The set of vectors is aggregated into a vector through a max pooling layer as the final image embedding.
	A GRU is used to encode each caption into a text embedding of dimension 1024.
	Our reproduced VSE++$ ^{\dagger} $ has superior performance compared to the original VSE++~\cite{faghri2018vse++}.
	VSE++$ ^{\dagger} $ is trained using Adam~\cite{kingma2015adam} for 20 epochs, with a batch size of 128 for both datasets. 
	The learning rate of the model is set as 0.0005. 
	\item 
	\textbf{VSE$ \infty $} is the current state-of-the-art global-level alignment model.
	VSE$ \infty $ introduces the Generalized Pooling Operator~(GPO) for aggregating image and text features.
	GPO learns the best pooling strategy for different features to generate global features.
	VSE$ \infty $ uses BUTD attention region features as image features. 
	The text encoder uses the Bidirectional Encoder Representations from Transformers~(BERT).
	VSE$ \infty $ are trained using AdamW~\cite{loshchilov2018decoupled} for 30 epochs, with a batch size of 128.
	The initial learning rate of the model is set as 0.0005 for the first 15 epochs and then decays by a factor of 10 for the last 15 epochs. 
	\item 
	\textbf{SGRAF} is a Similarity Graph Reasoning and Attention Filtration network for ITR.
	SGRAF is a classic local-level alignment model, which consists of two models, Similarity Graph Reasoning~(\textbf{SGR}) model and Similarity Attention Filtration~(\textbf{SAF}) model
	We report the performance of the two models individually as well as the performance of the ensemble model.
	SGRAF is trained using Adam for 20 epochs, with a batch size of 128 for both datasets. 
	The initial learning rate of the model is set as 0.0005 for first 10 epochs, and then decays by a factor of 10 for the last 10 epochs.  
\end{itemize}
All baseline models use the most commonly used pairwise loss, triplet loss $ \mathcal{L}_{\text{Triplet}} $, as the optimization objective.
Hyperparameters are set to $ \lambda = 0.2 $ and $ \tau = 10^{-2} $ for both datasets.

\subsubsection{Evaluation Metric}
For the evaluation on the Flickr30K and MS-COCO dataset, we use Recall@K~(R@K), with $ K = \{1,5,10\} $ as the evaluation metric. 
R@K indicates the percentage of queries for which the model returns the correct item in its top $ K $ results. 
We use RSUM, which is defined as the sum of recall metrics at $ K = \{1,5,10\} $ of both text-to-image and image-to-text retrievals, as an average metric to gauge the overall retrieval performance.
In order to verify the quality of relevance ranking, we conduct experiments on the ECCV Caption dataset.
Besides R@K, following Chun~\textit{et al.}~\cite{chun2022eccv}, we use mAP@R and R-P as the evaluation metrics. 

\subsection{Improvements on Existing ITR Models}
To verify that listwise ranking can improve the retrieval performance of current conventional pairwise-based ITR models, we conduct experiments on the VSE++~\cite{faghri2018vse++}, VSE$ \infty $~\cite{chen2021learning}, and SGRAF~\cite{diao2021similarity} models.
Our listwise ranking approach can be plug-and-play applied to these models.
Therefore, we integrate listwise ranking into these models, denoted as \textbf{+ \textit{Listwise}}.
\tablename~\ref{f30k} shows the improvements on these models on the Flickr30K dataset.
For a fair comparison, the models we compare all use the BUTD features~\cite{anderson2018bottom} as image features.
Integrating listwise ranking improves the retrieval performance of the three baseline models.
For the local-level alignment model SGRAF, the single-model SAF and SGR improve by 9.8\% and 10.6\% on RSUM, respectively.
The ensemble model SGRAF also achieves a significant improvement of 9.5\% on RSUM.
For the global level alignment models VSE++ and VSE$ \infty $, the RSUM is improved by 5.0\% and 3.9\% after integrating the listwise approach, respectively.
\textbf{*} represents the retrieval result of the ensemble model of two single models.
The VSE$ \infty $ model achieves state-of-the-art performance after integrating listwise ranking.

\tablename~\ref{coco1k} and \tablename~\ref{coco5k} show the improvements on the MS-COCO 1K and 5K test sets.
After integrating listwise ranking, the three baselines improve on most of the evaluation metrics.
VSE$ \infty $ \textbf{+ \textit{Listwise*}} also achieves state-of-the-art performance on both test sets.
Extensive experiments on two datasets and three baseline models demonstrate the generality of our listwise ranking approach.
Listwise ranking can be flexibly integrated into current conventional pairwise-based ITR models to improve retrieval performance.

\begin{table}[t] 
	\caption{Ablation studies on the Flickr30K dataset}
	\setlength\tabcolsep{3.4pt}
	\begin{center}
		\begin{tabular}{ccccccccccccc}
			\toprule[1pt]
			\multirow{2}*{$ \mathcal{L}_{\text{Triplet}} $} & \multirow{2}*{$ \mathcal{L}_{\text{S-NDCG}} $} & \multicolumn{3}{c}{Image-to-Text} & & \multicolumn{3}{c}{Text-to-Image} & \multirow{2}*{RSUM}\\
			\cline{3-5}\cline{7-9}
			\specialrule{0em}{2pt}{0pt}
			~ & ~ & R@1 & R@5 & R@10 & & R@1 & R@5 & R@10 \\
			\hline
			
			\specialrule{0em}{2pt}{0pt}
			\checkmark & ~ & 70.4 & 90.7 & 95.2 & & 51.6 & 78.6 & 86.2 & 472.7 \\
			~ & \checkmark & 59.1 & 84.0 & 89.9 & & 39.5 & 69.8 & 79.7 & 422.0 \\
			\rowcolor{black!10}
			\checkmark & \checkmark & 70.6 & 91.9 & 95.6 & & 53.0 & 79.9 & 86.8 & \textbf{477.7} \\
			
			\bottomrule[1pt]
		\end{tabular}
	\end{center}
	\label{ablation_f30k}
\end{table}
\begin{table}[t] 
	\caption{Ablation studies on the MS-COCO dataset}
	\setlength\tabcolsep{3.4pt}
	\begin{center}
		\begin{tabular}{ccccccccccccc}
			\toprule[1pt]
			\multirow{2}*{$ \mathcal{L}_{\text{Triplet}} $} & \multirow{2}*{$ \mathcal{L}_{\text{S-NDCG}} $} & \multicolumn{3}{c}{Image-to-Text} & & \multicolumn{3}{c}{Text-to-Image} & \multirow{2}*{RSUM}\\
			\cline{3-5}\cline{7-9}
			\specialrule{0em}{2pt}{0pt}
			~ & ~ & R@1 & R@5 & R@10 & & R@1 & R@5 & R@10 \\
			\hline
			
			\specialrule{0em}{2pt}{0pt}
			\multicolumn{10}{c}{1K Test Set} \\
			\hline
			
			\specialrule{0em}{2pt}{0pt}
			\checkmark & ~ & 72.8 & 94.6 & 98.1 & & 57.9 & 88.0 & 94.4 & 505.8 \\
			~ & \checkmark & 62.1 & 88.8 & 95.3 & & 49.0 & 82.1 & 90.9 & 468.2 \\
			\rowcolor{black!10}
			\checkmark & \checkmark & 73.4 & 94.4 & 98.1 & & 58.8 & 88.4 & 94.6 & \textbf{507.7} \\
			\hline
			
			\specialrule{0em}{2pt}{0pt}
			\multicolumn{10}{c}{5K Test Set} \\
			\hline
			
			\specialrule{0em}{2pt}{0pt}
			\checkmark & ~ & 49.9 & 78.4 & 88.0 & & 35.6 & 65.9 & 77.5 & 395.4 \\
			~ & \checkmark & 37.0 & 66.7 & 78.5 & & 27.9 & 56.1 & 69.2 & 335.3 \\
			\rowcolor{black!10}
			\checkmark & \checkmark & 50.5 & 79.5 & 88.1 & & 36.0 & 66.4 & 78.2 & \textbf{398.6} \\
			
			\bottomrule[1pt]
		\end{tabular}
	\end{center}
	\label{ablation_coco}
\end{table}
\begin{table}[t] 
	\caption{Ablation studies on the ECCV Caption dataset}
	\setlength\tabcolsep{3.8pt}
	\begin{center}
		\begin{tabular}{ccccccccccccc}
			\toprule[1pt]
			\multirow{2}*{$ \mathcal{L}_{\text{Triplet}} $} & \multirow{2}*{$ \mathcal{L}_{\text{S-NDCG}} $} &
			\multicolumn{3}{c}{Image-to-Text} & & \multicolumn{3}{c}{Text-to-Image} \\
			\cline{3-5}\cline{7-9}
			\specialrule{0em}{2pt}{0pt}
			~ & ~ & mAP@R & R-P & R@1 & & mAP@R & R-P & R@1 \\
			\hline
			
			\specialrule{0em}{2pt}{0pt}
			\checkmark & ~ & 26.80 & 38.98 & 65.50 & & 46.29 & 54.72 & 85.44 \\
			~ & \checkmark & 21.40 & 33.53 & 56.38 & & 41.72 & 50.72 & 80.56 \\
			\rowcolor{black!10}
			\checkmark & \checkmark & \textbf{27.75} & \textbf{39.68} & \textbf{68.83} & & \textbf{47.35} & \textbf{55.56} & \textbf{86.49} \\
			
			\bottomrule[1pt]
		\end{tabular}
	\end{center}
	\label{ablation_eccv}
\end{table}
\subsection{Ablation Studies}
\subsubsection{Impact of Different Modules}
This paper proposes to integrate listwise ranking into conventional pairwise-based ITR models.
We experimentally verify the impact of different modules using the VSE++$ ^{\dag} $ model as the baseline.
\tablename~\ref{ablation_f30k} shows the results of ablation studies on the Flickr30K dataset.
Using $ \mathcal{L}_{\text{S-NDCG}} $ alone does not achieve considerable retrieval performance.
This is because NDCG cannot be globally optimized for end-to-end training.
Training a deep learning model requires dividing the data into batches.
There is a gap between the NDCG on each batch and the NDCG on the entire dataset.
Optimizing $ \mathcal{L}_{\text{S-NDCG}} $ alone within a batch does not generalize well to the entire dataset.
$ \mathcal{L}_{\text{Triplet}} $ has better generalization, but it cannot rank all samples by relevance.
As shown in \tablename~\ref{ablation_f30k}, the best performance can be achieved by combining the $ \mathcal{L}_{\text{Triplet}} $ and $ \mathcal{L}_{\text{S-NDCG}} $.
Integrating listwise ranking into the pairwise-based ITR model can not only guarantee generalization, but also achieve the relevance ranking of all candidates.
\tablename~\ref{ablation_coco} shows the ablation studies on MS-COCO 1K and 5K test sets.
Experimental results on these two test sets are consistent with those on the Flickr30K dataset.
$ \mathcal{L}_{\text{Triplet}} $ and $ \mathcal{L}_{\text{S-NDCG}} $ can complement each other's defects.

The evaluation metric R@K on the Flickr30K and MS-COCO datasets can only reflect the ranking performance of positive samples, but cannot reflect the quality of all samples ranked by relevance.
To this end, we conduct an ablation study on the ECCV Caption dataset.
As shown in \tablename~\ref{ablation_eccv}, using $ \mathcal{L}_{\text{Triplet}} $ and $ \mathcal{L}_{\text{S-NDCG}} $ together has a significant performance improvement compared to using either loss alone.
This shows that integrating listwise ranking into the pairwise-based ITR model can achieve better relevance ranking quality.

\begin{figure}[t]
	\centering
	\subfigure[Losses during training.]{
		\label{train_loss}
		\includegraphics[width=0.465\linewidth]{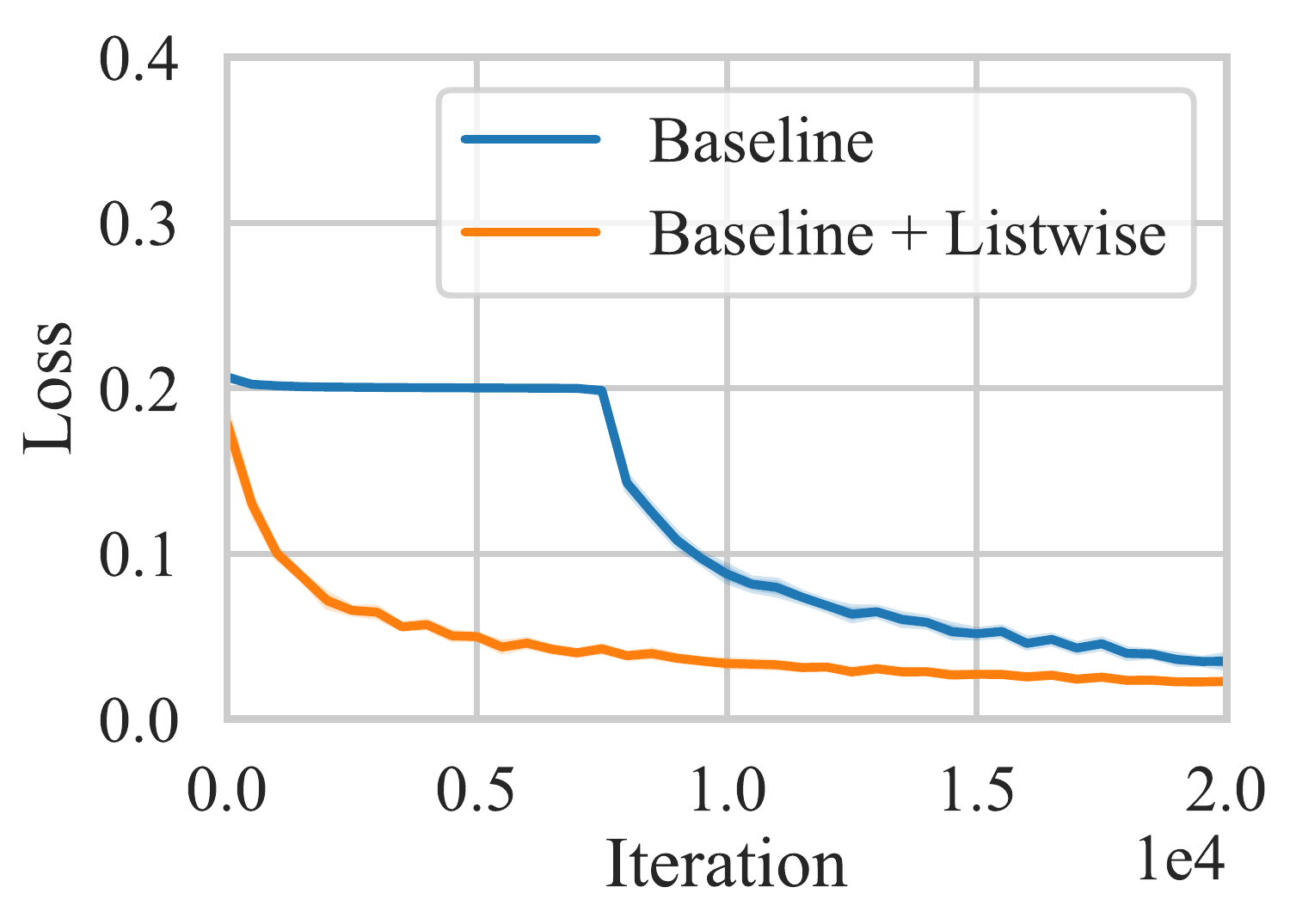}
	}
	\subfigure[RSUM during training.]{
		\label{train_rsum}
		\includegraphics[width=0.465\linewidth]{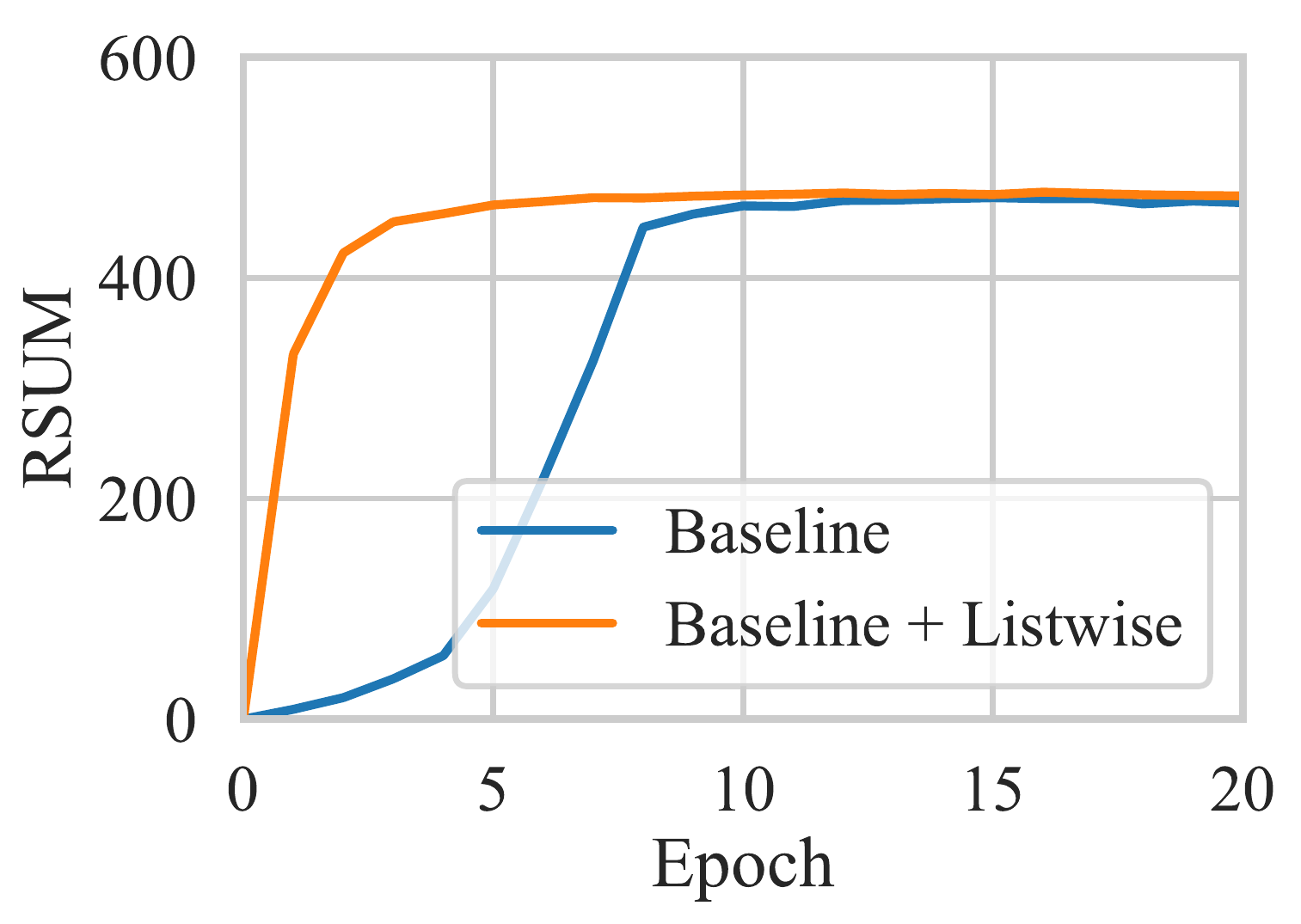}
	}
	\caption{Training behavior.}
	\label{convergence}
\end{figure}
\begin{figure}[t]
	\centering
	\includegraphics[width=0.7\linewidth]{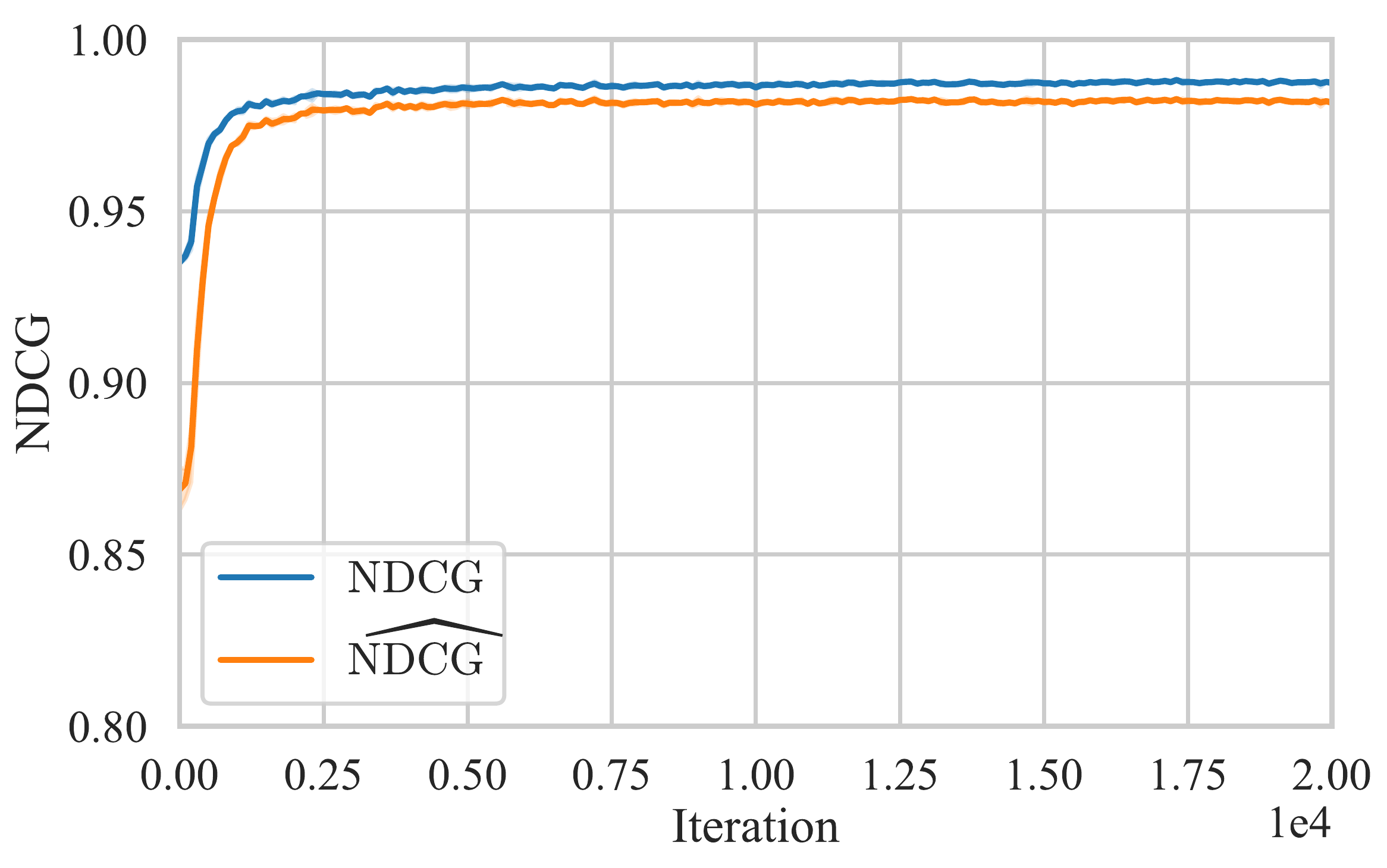}
	\caption{NDCG and $ \widehat{\text{NDCG}} $ on each batch during training.}
	\label{train_ndcg}
\end{figure}
\subsubsection{Training Behavior}
We compare the training behaviors of the conventional pairwise-based ITR model and the model integrating listwise ranking.
The experiment is conducted on the Flickr30K dataset.
We adopt the VSE++$ ^{\dagger} $ model as the baseline.
\figurename~\ref{train_loss} shows the losses during training.
The loss function of the baseline model $ \mathcal{L}_{\text{Triplet}} $ converges slowly.
After integrating listwise ranking, the loss function converges quickly.
The pairwise approach generally adopts a hard negative mining strategy to improve performance.
$ \mathcal{L}_{\text{Triplet}} $ uses the hardest negative samples mined for training, which have been shown to lead to bad local minima in the early phases of the optimization~\cite{xuan2020hard}.
Instead of optimizing on individual sample pairs, $ \mathcal{L}_{\text{S-NDCG}} $ optimizes on the entire ranking list.
Incorporating more samples from the ranking list into the training effectively avoids the local minima problem and improves the convergence of the training.
\figurename~\ref{train_rsum} shows the RSUM on the validation set during training.
Integrating listwise ranking can not only achieve the highest retrieval performance but also converge quickly.

\figurename~\ref{train_ndcg} shows the NDCG and $ \widehat{\text{NDCG}} $ on each batch during training.
The NDCG in \figurename~\ref{train_ndcg} is the real NDCG calculated by using each batch as a ranking list.
$ \widehat{\text{NDCG}} $ is the NDCG approximated by our approach.
NDCG and $ \widehat{\text{NDCG}} $ have the same tendency during training.
Our listwise ranking approach can improve the real NDCG by optimizing $ \widehat{\text{NDCG}} $.

\begin{figure}[t]
	\centering
	\includegraphics[width=0.7\linewidth]{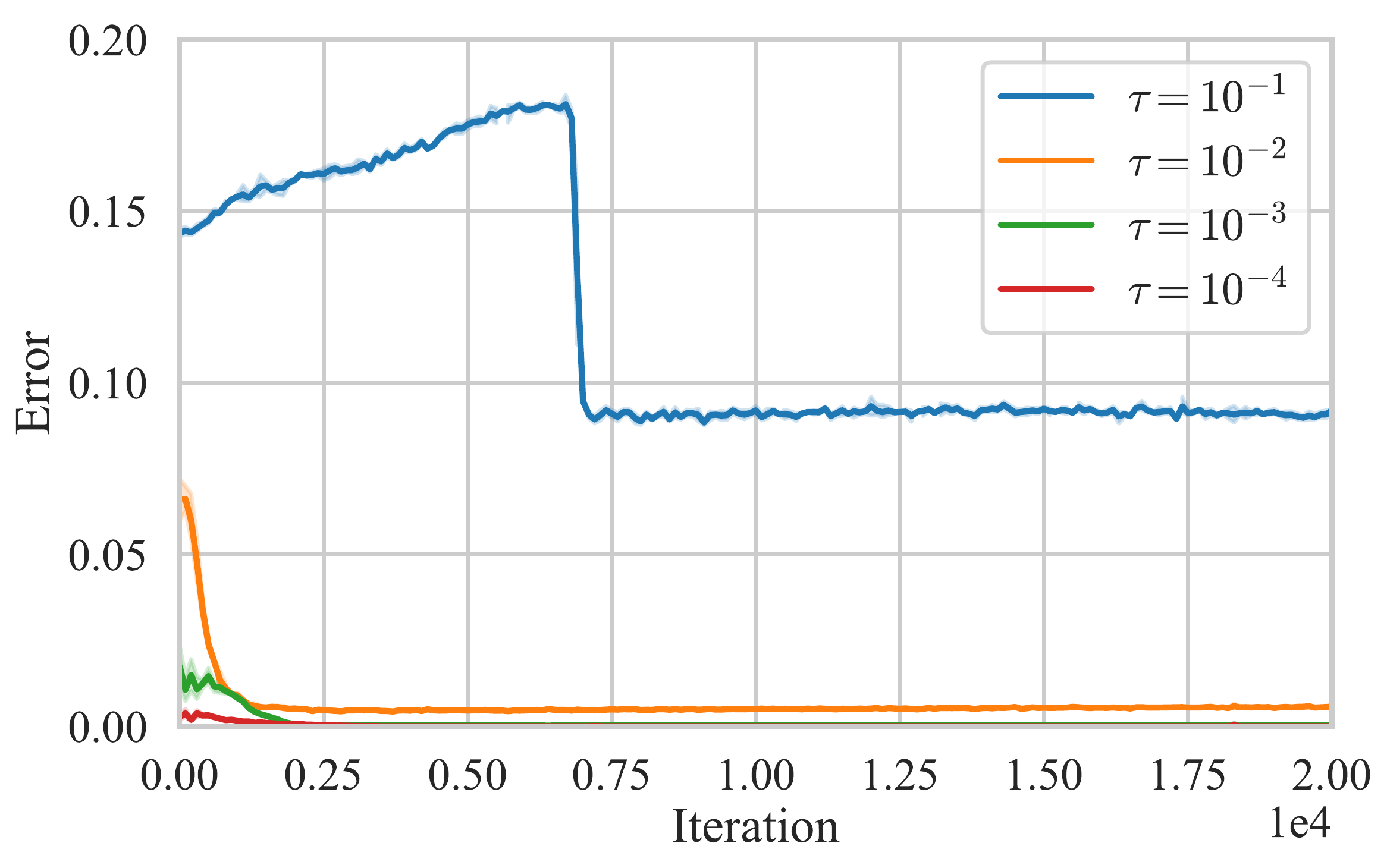}
	\caption{NDCG approximation error during training.}
	\label{error}
\end{figure}
\subsubsection{NDCG Approximation Error}
Since $ \widehat{\text{NDCG}} $ is a smooth approximation of NDCG, we need to evaluate the approximate error between $ \widehat{\text{NDCG}} $ and NDCG.
The NDCG approximation error is defined as:
\begin{equation}
	\text{Error}
	= 
	\left|
	\widehat{\text{NDCG}}
	- {\text{NDCG}}
	\right|.
\end{equation}
\figurename~\ref{error} shows the approximation error for different $ \tau $ during training.
The experiment is conducted on the Flickr30K dataset.
We adopt the VSE++$ ^{\dagger} $ model as the baseline.
It can be seen from \figurename~\ref{error} that the smaller $ \tau $ is, the smaller the error is.
In the initial stage of training, the error is large.
As the training progresses, the error can decrease to a stable level.
When $ \tau < 10^{-2} $, the approximation error can be stabilized below 0.01.
This shows that our approach can accurately approximate NDCG when the value of $ \tau $ is appropriate.

\begin{figure}[t]
	\centering
	\subfigure[Flickr30K]{
		\label{tau_rsum_f30k}
		\includegraphics[width=0.465\linewidth]{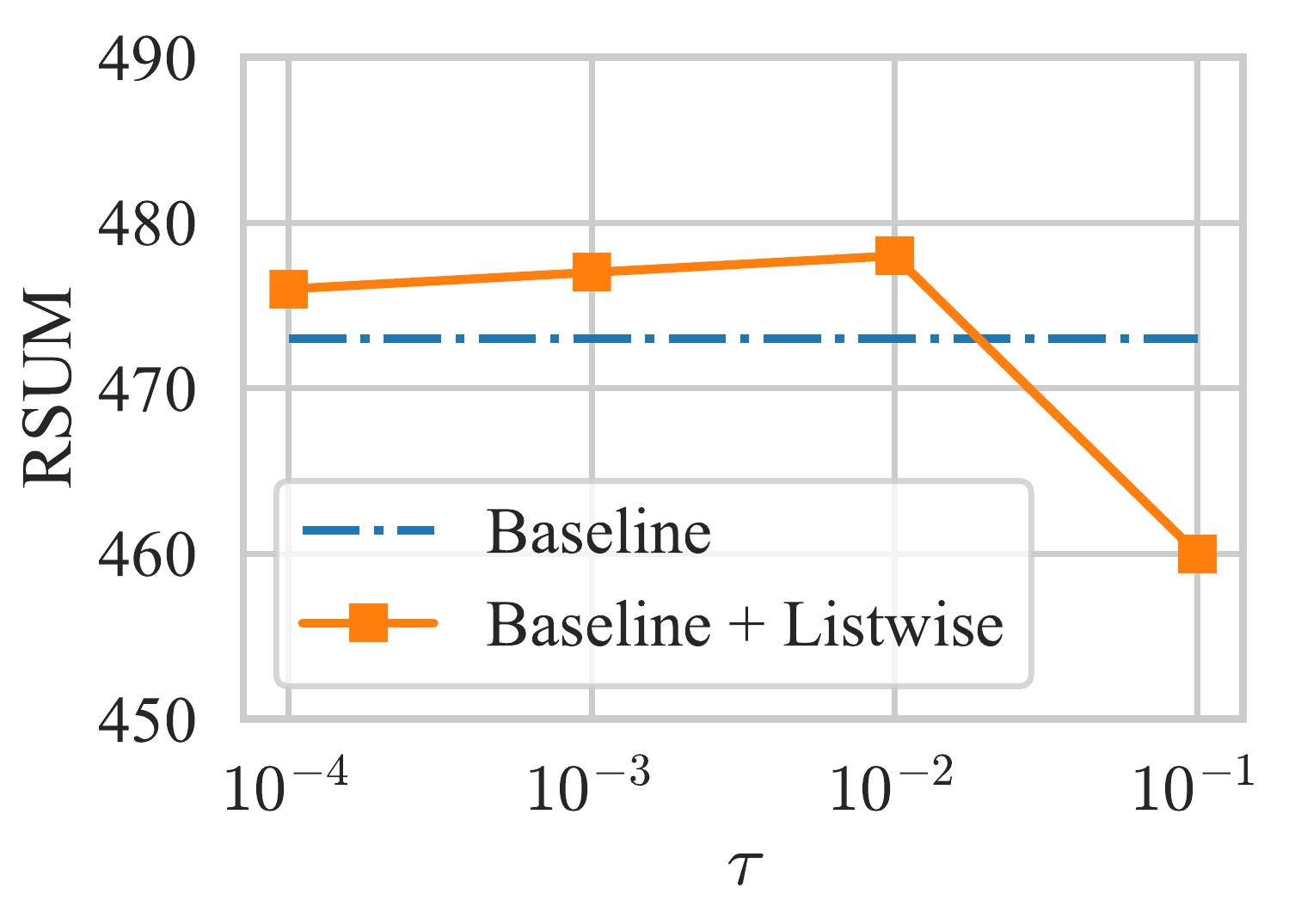}
	}
	\subfigure[MS-COCO 5K]{
		\label{tau_rsum_coco}
		\includegraphics[width=0.465\linewidth]{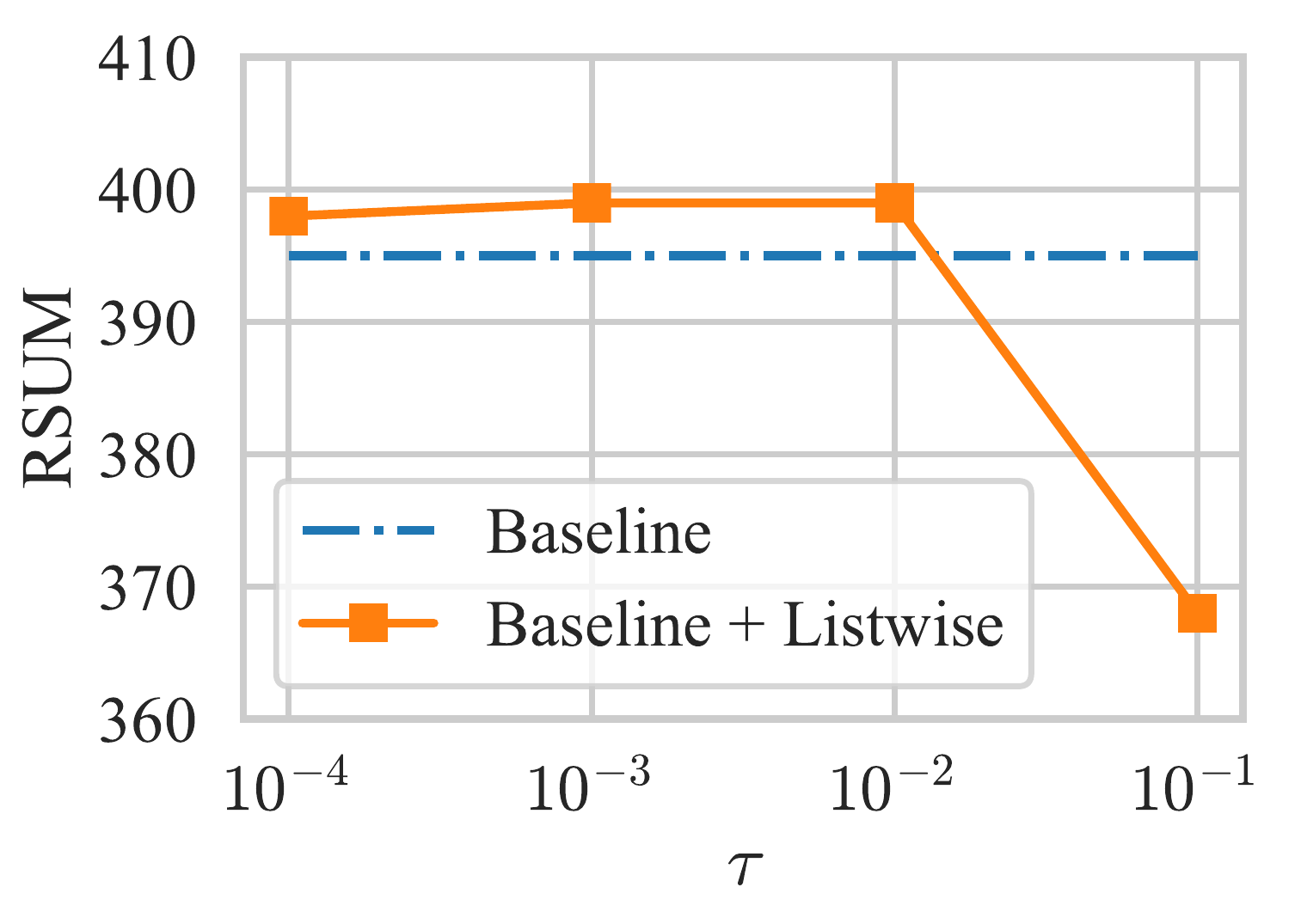}
	}
	\caption{Impact of $ \tau $  on the Flickr30K and MS-COCO dataset.}
	\label{tau_rsum}
\end{figure}
\begin{figure}[t]
	\centering
	\subfigure[Image-to-Text]{
		\label{tau_map_i2t}
		\includegraphics[width=0.465\linewidth]{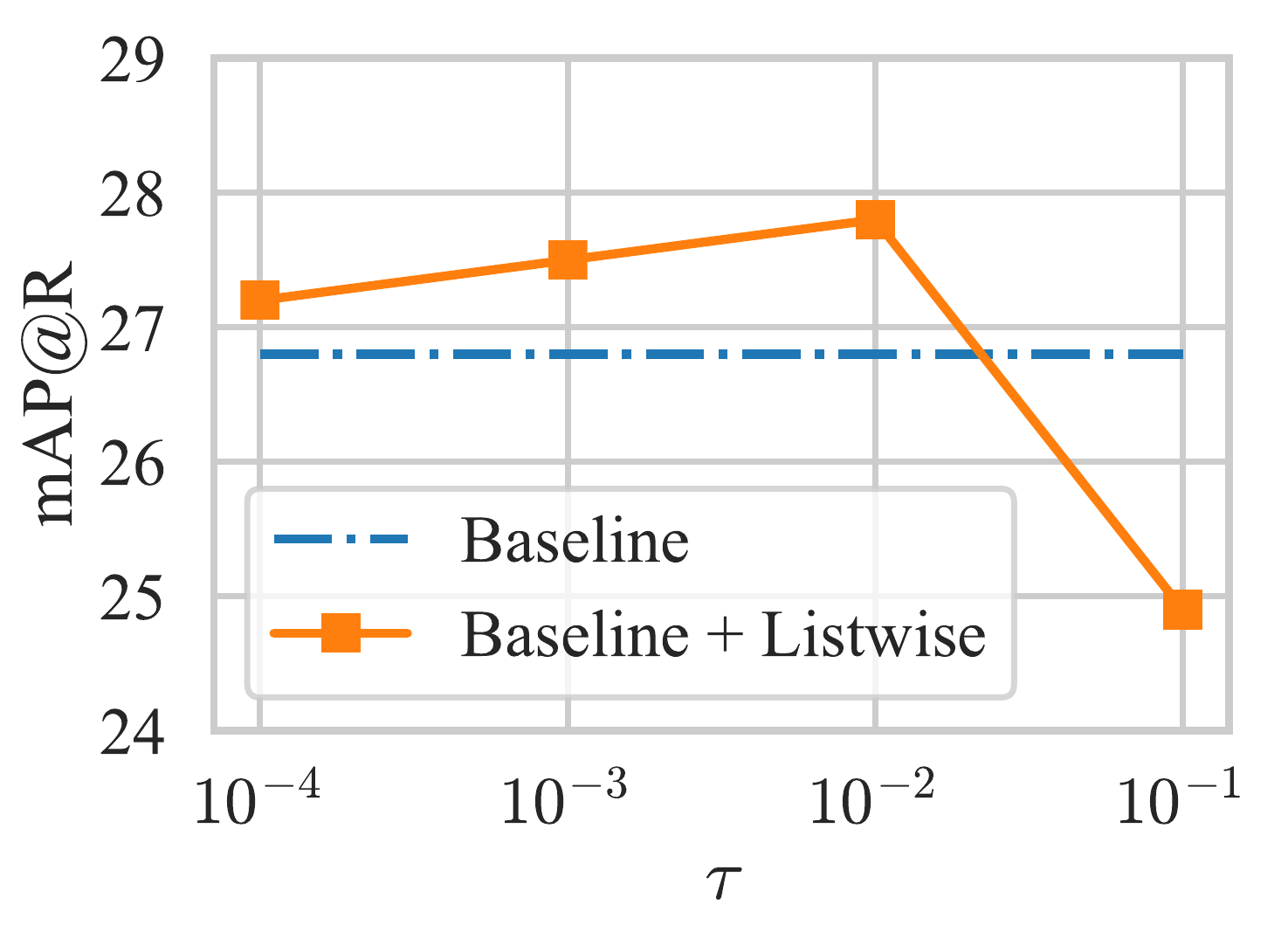}
	}
	\subfigure[Text-to-Image]{
		\label{tau_map_t2i}
		\includegraphics[width=0.465\linewidth]{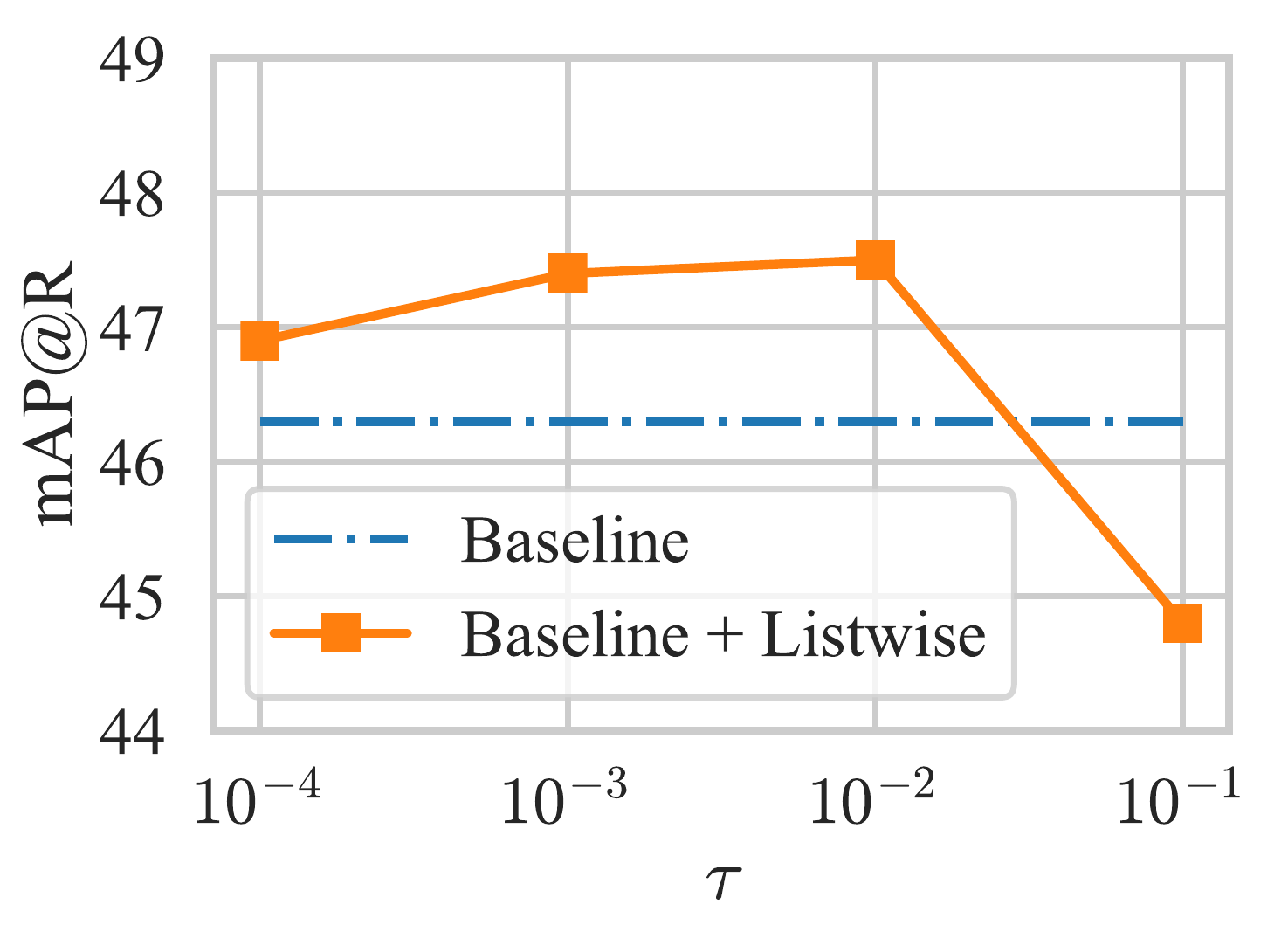}
	}
	\caption{Impact of $ \tau $  on the ECCV Caption dataset.}
	\label{tau_map}
\end{figure}
\subsubsection{Hyperparameter Analysis}
There is one hyperparameter temperature coefficient $ \tau $ in our listwise ranking approach that can be tuned.
We experiment with several parameter settings on the three datasets using the VSE++$ ^{\dagger} $ model as the baseline.
\figurename~\ref{tau_rsum_f30k} shows the impact of $ \tau $  on the Flickr30K dataset.
When $ \tau < 10^{-2} $, the retrieval performance of the model integrated with listwise ranking is always better than the baseline.
In addition, when $ \tau < 10^{-2} $, there is no obvious change in retrieval performance.
This shows that our proposed listwise ranking approach is not sensitive to the hyperparameter, and is convenient to integrate into various ITR models.
In particular, when $ \tau = 10^{-1} $, the retrieval performance decreases significantly.
It can be seen from \figurename~\ref{error} that the approximation error is large when $ \tau = 10^{-1} $.
Therefore, the large approximation error leads to a decrease in retrieval performance.
\figurename~\ref{tau_rsum_coco} shows the impact of $ \tau $  on the MS-COCO 5K test set.
The experimental results are consistent with \figurename~\ref{tau_rsum_f30k}.

\figurename~\ref{tau_map} shows the impact of the hyperparameter $ \tau $ on the ECCV Caption dataset.
The retrieval performance on this dataset can reflect the quality of relevance ranking.
Similar to \figurename~\ref{tau_rsum}, when $ \tau = 10^{-1} $, the retrieval performance decreases significantly.
The reason is that the approximation error at $ \tau = 10^{-1} $ is large.
In particular, when $ \tau = 10^{-2} $, the retrieval performance is the best. 
When $ \tau < 10^{-2} $, the retrieval performance decreases slightly.
This is because the smaller the $ \tau $, the smaller the activation area that can provide gradients for ranking optimization.
Small activation areas can negatively affect ranking optimization.
Therefore, when $ \tau < 10^{-2} $, the performance of the relevance ranking decreases.
There is a trade-off between the size of the activation area and the approximation error.
When $ \tau = 10^{-2} $, a smaller approximation error and a larger activation area can be satisfied at the same time.

\begin{table}[t]
	\caption{Computational efficiency on the Flickr30K dataset.}
	\setlength\tabcolsep{12pt}
	\begin{center}
		\begin{tabular}{lcccccccccccc}
			\toprule[1pt]
			Model & Training Time(s) per Epoch & Testing Time(s) \\
			\hline
			\specialrule{0em}{2pt}{0pt}
			VSE$ \infty $ & \textbf{229.6 $ \pm $ 5.6} & \textbf{4.6 $ \pm $ 0.1} \\
			\textbf{+ \textit{Listwise}} & 241.9 $ \pm $ 5.4 & \textbf{4.6 $ \pm $ 0.1} \\
			\bottomrule[1pt]
		\end{tabular}
	\end{center}
	\label{efficiency}
\end{table}
\subsection{Computational Efficiency}
Our listwise ranking approach can be plug-and-play applied to current pairwise-based ITR models, which slightly increases the computational overhead of training but not inference.
We evaluate the computational efficiency of listwise ranking on the Flickr30K dataset.
We adopt the VSE$ \infty $ model as the baseline.
We integrate listwise ranking into the baseline model, denoted as \textbf{+ \textit{Listwise}}.
Our computational efficiency experiments are all performed on an NVIDIA GeForce RTX 3090 GPU.
All experimental conditions are the same for both models during training and testing.
The experimental results are shown in \tablename~\ref{efficiency}.

\subsubsection{Training Phase}
We count the time it takes for these two models to train an epoch, respectively.
\tablename~\ref{efficiency} shows that integrating listwise ranking increases the training time by 5\%.
As analyzed in Subsection~\ref{overhead}, the RSC module calculates the relevance score, and the approximate calculation of the NDCG can increase the complexity of training.
Due to our optimization in engineering implementation, integrating listwise ranking only adds a small amount of training time.

\subsubsection{Testing Phase}
We count the time spent in the testing phase on the Flickr30K dataset.
\tablename~\ref{efficiency} shows that integrating listwise ranking takes the same time as the baseline in the testing phase.
The computation of relevance scores and the approximate NDCG are no longer required during the testing phase.
The testing efficiency of integrating listwise ranking is the same as the original model.

\subsection{Qualitative Results}
\begin{figure*}[t]
	\centering
	\includegraphics[width=\linewidth]{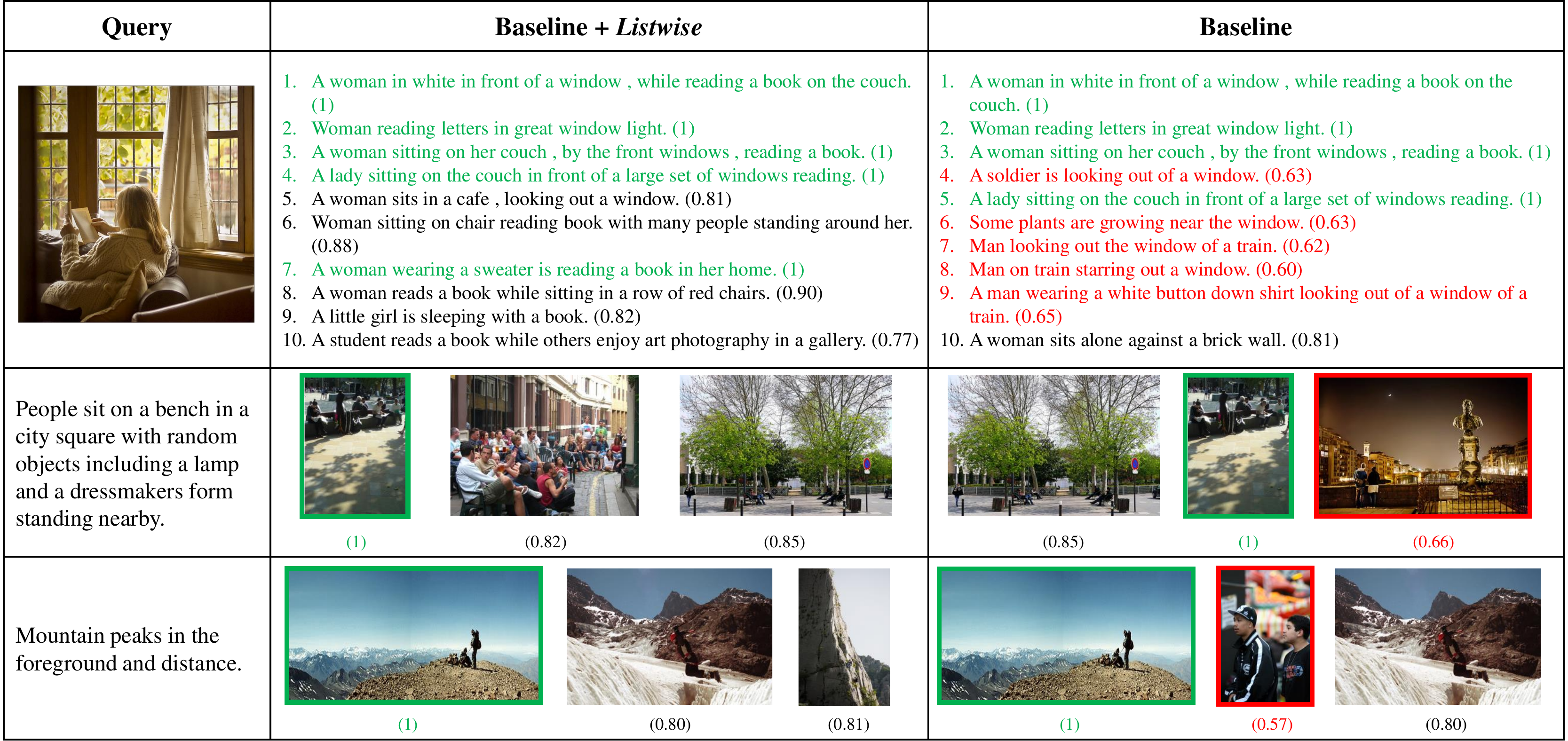}
	\caption{Qualitative comparisons between the conventional pairwise-based ITR model and the model integrating listwise ranking.}
	\label{qualitative}
\end{figure*}
\figurename~\ref{qualitative} shows the qualitative comparisons between the conventional pairwise-based ITR model and the model integrating listwise ranking.
The experiment is conducted on the Flickr30K dataset.
We adopt the VSE++$ ^{\dagger} $ model as the baseline.
For image-to-text retrieval, given an image query, we show the Top-10 retrieved captions.
For text-to-image retrieval, given a caption query, we show the Top-3 retrieved images.
The number in brackets is the relevance score between the query and the candidates.
The ground truth retrieval items for each query are outlined in green.
Obviously wrong retrieval results are marked in red.
As shown in \figurename~\ref{qualitative}, the relevance score between the results retrieved by the model integrating listwise ranking and the query is higher than the baseline.
Overall, listwise ranking improves the relevance ranking performance of retrieval results.
The incorrect retrieval items are semantically closer to the query compared with the baseline.
For example, in the third row in \figurename~\ref{qualitative}, the incorrect results retrieved are also related to ``mountains''.
This benefits from the proposed listwise loss $ \mathcal{L}_{\text{S-NDCG}} $, which is an approximate function of NDCG.
Our listwise ranking approach optimizes the ranking list, so that candidates with higher relevance scores are mapped closer to the query than those with lower relevance scores.
In contrast, some obviously wrong results appear in the retrieval results of the baseline.
This is because the pairwise-based ITR model is optimized based on the original pairwise annotations.
The pairwise loss $ \mathcal{L}_{\text{Triplet}} $ only focuses on the correct ranking of positive samples, and cannot rank negative samples by relevance.
Qualitative experimental results illustrate that integrating listwise ranking can provide more user-friendly retrieval results.

\section{Conclusion}
This paper integrates listwise ranking into pairwise-based Image-Text Retrieval~(ITR).
Specifically, we propose a Relevance Score Calculation~(RSC) module to calculate the relevance score of the entire ranked list.
Then we transform the non-differentiable ranking metric NDCG into a differentiable listwise loss, named Smooth-NDCG~(S-NDCG).
Our listwise ranking approach can be plug-and-play integrated into conventional pairwise-based ITR models to improve performance.
More importantly, integrating listwise ranking can provide more user-friendly retrieval results.
In future work, we plan to expand listwise ranking to more cross-modal retrieval tasks.

%\clearpage
%\balance
\bibliography{ref}

\begin{thebibliography}{10}
\providecommand{\url}[1]{#1}
\csname url@samestyle\endcsname
\providecommand{\newblock}{\relax}
\providecommand{\bibinfo}[2]{#2}
\providecommand{\BIBentrySTDinterwordspacing}{\spaceskip=0pt\relax}
\providecommand{\BIBentryALTinterwordstretchfactor}{4}
\providecommand{\BIBentryALTinterwordspacing}{\spaceskip=\fontdimen2\font plus
\BIBentryALTinterwordstretchfactor\fontdimen3\font minus
  \fontdimen4\font\relax}
\providecommand{\BIBforeignlanguage}[2]{{%
\expandafter\ifx\csname l@#1\endcsname\relax
\typeout{** WARNING: IEEEtran.bst: No hyphenation pattern has been}%
\typeout{** loaded for the language `#1'. Using the pattern for}%
\typeout{** the default language instead.}%
\else
\language=\csname l@#1\endcsname
\fi
#2}}
\providecommand{\BIBdecl}{\relax}
\BIBdecl

\bibitem{zhang2020context}
Q.~Zhang, Z.~Lei, Z.~Zhang, and S.~Z. Li, ``Context-aware attention network for
  image-text retrieval,'' in \emph{Proc. {IEEE/CVF} Conf. Comput. Vis. and
  Pattern Recognit.~({CVPR})}, 2020, pp. 3536--3545.

\bibitem{chen2020imram}
H.~Chen, G.~Ding, X.~Liu, Z.~Lin, J.~Liu, and J.~Han, ``Imram: Iterative
  matching with recurrent attention memory for cross-modal image-text
  retrieval,'' in \emph{Proc. {IEEE/CVF} Conf. Comput. Vis. and Pattern
  Recognit.~({CVPR})}, 2020, pp. 12\,655--12\,663.

\bibitem{qu2021dynamic}
L.~Qu, M.~Liu, J.~Wu, Z.~Gao, and L.~Nie, ``Dynamic modality interaction
  modeling for image-text retrieval,'' in \emph{Proc. Int. ACM SIGIR Conf. Res.
  Develop. Inf. Retr.~({SIGIR})}, 2021, pp. 1104--1113.

\bibitem{wei2023less}
J.~Wei, Y.~Yang, X.~Xu, J.~Song, G.~Wang, and H.~T. Shen, ``Less is better:
  Exponential loss for cross-modal matching,'' \emph{{IEEE} Trans. Circuits
  Syst. Video Technol.}, 2023.

\bibitem{zhu2023esa}
H.~Zhu, C.~Zhang, Y.~Wei, S.~Huang, and Y.~Zhao, ``Esa: External space
  attention aggregation for image-text retrieval,'' \emph{{IEEE} Trans.
  Circuits Syst. Video Technol.}, 2023.

\bibitem{dong2022hierarchical}
X.~Dong, H.~Zhang, L.~Zhu, L.~Nie, and L.~Liu, ``Hierarchical feature
  aggregation based on transformer for image-text matching,'' \emph{{IEEE}
  Trans. Circuits Syst. Video Technol.}, vol.~32, no.~9, pp. 6437--6447, 2022.

\bibitem{li2023integrating}
Z.~Li, C.~Guo, Z.~Feng, J.-N. Hwang, and Z.~Du, ``Integrating language guidance
  into image-text matching for correcting false negatives,'' \emph{IEEE
  Transactions on Multimedia}, 2023.

\bibitem{schroff2015facenet}
F.~Schroff, D.~Kalenichenko, and J.~Philbin, ``Facenet: A unified embedding for
  face recognition and clustering,'' in \emph{Proc. {IEEE/CVF} Conf. Comput.
  Vis. and Pattern Recognit.~({CVPR})}, 2015, pp. 815--823.

\bibitem{lee2018stacked}
K.-H. Lee, X.~Chen, G.~Hua, H.~Hu, and X.~He, ``Stacked cross attention for
  image-text matching,'' in \emph{Proc. Eur. Conf. Comput. Vis.~({ECCV})},
  2018, pp. 201--216.

\bibitem{li2019visual}
K.~Li, Y.~Zhang, K.~Li, Y.~Li, and Y.~Fu, ``Visual semantic reasoning for
  image-text matching,'' in \emph{Proc. {IEEE/CVF} Int. Conf. Comput.
  Vis.~({ICCV})}, 2019, pp. 4654--4662.

\bibitem{liu2020graph}
C.~Liu, Z.~Mao, T.~Zhang, H.~Xie, B.~Wang, and Y.~Zhang, ``Graph structured
  network for image-text matching,'' in \emph{Proc. {IEEE/CVF} Conf. Comput.
  Vis. and Pattern Recognit.~({CVPR})}, 2020, pp. 10\,921--10\,930.

\bibitem{diao2021similarity}
H.~Diao, Y.~Zhang, L.~Ma, and H.~Lu, ``Similarity reasoning and filtration for
  image-text matching,'' in \emph{Proc. AAAI Conf. Artif. Intell.~({AAAI})},
  vol.~35, no.~2, 2021, pp. 1218--1226.

\bibitem{zhang2022negative}
K.~Zhang, Z.~Mao, Q.~Wang, and Y.~Zhang, ``Negative-aware attention framework
  for image-text matching,'' in \emph{Proc. {IEEE/CVF} Conf. Comput. Vis. and
  Pattern Recognit.~({CVPR})}, 2022, pp. 15\,661--15\,670.

\bibitem{faghri2018vse++}
F.~Faghri, D.~J. Fleet, J.~R. Kiros, and S.~Fidler, ``Vse++: Improving
  visual-semantic embeddings with hard negatives,'' in \emph{Proc. Brit. Mach.
  Vis. Conf.~({BMVC})}, 2018.

\bibitem{chen2020adaptive}
T.~Chen, J.~Deng, and J.~Luo, ``Adaptive offline quintuplet loss for image-text
  matching,'' in \emph{Proc. Eur. Conf. Comput. Vis.~({ECCV})}, 2020.

\bibitem{wei2020universal}
J.~Wei, X.~Xu, Y.~Yang, Y.~Ji, Z.~Wang, and H.~T. Shen, ``Universal weighting
  metric learning for cross-modal matching,'' in \emph{Proc. {IEEE/CVF} Conf.
  Comput. Vis. and Pattern Recognit.~({CVPR})}, 2020.

\bibitem{wei2021universal}
J.~Wei, Y.~Yang, X.~Xu, X.~Zhu, and H.~T. Shen, ``Universal weighting metric
  learning for cross-modal retrieval,'' \emph{{IEEE} Trans. Pattern Anal. Mach.
  Intell.}, 2021.

\bibitem{wei2021meta}
J.~Wei, X.~Xu, Z.~Wang, and G.~Wang, ``Meta self-paced learning for cross-modal
  matching,'' in \emph{Proc. ACM Int. Conf. Multimedia~({ACM MM})}, 2021.

\bibitem{chun2022eccv}
S.~Chun, W.~Kim, S.~Park, M.~Chang, and S.~J. Oh, ``Eccv caption: Correcting
  false negatives by collecting machine-and-human-verified image-caption
  associations for ms-coco,'' in \emph{Proc. Eur. Conf. Comput.
  Vis.~({ECCV})}.\hskip 1em plus 0.5em minus 0.4em\relax Springer, 2022, pp.
  1--19.

\bibitem{cakir2019deep}
F.~Cakir, K.~He, X.~Xia, B.~Kulis, and S.~Sclaroff, ``Deep metric learning to
  rank,'' in \emph{Proc. {IEEE/CVF} Conf. Comput. Vis. and Pattern
  Recognit.~({CVPR})}, 2019, pp. 1861--1870.

\bibitem{brown2020smooth}
A.~Brown, W.~Xie, V.~Kalogeiton, and A.~Zisserman, ``Smooth-ap: Smoothing the
  path towards large-scale image retrieval,'' in \emph{Proc. Eur. Conf. Comput.
  Vis.~({ECCV})}.\hskip 1em plus 0.5em minus 0.4em\relax Springer, 2020, pp.
  677--694.

\bibitem{wang2020consensus}
H.~Wang, Y.~Zhang, Z.~Ji, Y.~Pang, and L.~Ma, ``Consensus-aware visual-semantic
  embedding for image-text matching,'' in \emph{Proc. Eur. Conf. Comput.
  Vis.~({ECCV})}.\hskip 1em plus 0.5em minus 0.4em\relax Springer, 2020, pp.
  18--34.

\bibitem{chen2021learning}
J.~Chen, H.~Hu, H.~Wu, Y.~Jiang, and C.~Wang, ``Learning the best pooling
  strategy for visual semantic embedding,'' in \emph{Proc. {IEEE/CVF} Conf.
  Comput. Vis. and Pattern Recognit.~({CVPR})}, 2021, pp. 15\,789--15\,798.

\bibitem{li2022multi}
Z.~Li, C.~Guo, Z.~Feng, J.-N. Hwang, and X.~Xue, ``Multi-view visual semantic
  embedding,'' in \emph{Proc. Int. Joint Conf. Artif. Intell.~({IJCAI})}, 2022.

\bibitem{frome2013devise}
A.~Frome, G.~S. Corrado, J.~Shlens, S.~Bengio, J.~Dean, M.~Ranzato, and
  T.~Mikolov, ``Devise: A deep visual-semantic embedding model,'' in
  \emph{Proc. Adv. Neural Inf. Process. Syst.~({NeurIPS})}, vol.~26, 2013.

\bibitem{chakrabarti2008structured}
S.~Chakrabarti, R.~Khanna, U.~Sawant, and C.~Bhattacharyya, ``Structured
  learning for non-smooth ranking losses,'' in \emph{Proc. ACM Int. Conf.
  Knowl. Discov. Data Mining~({SIGKDD})}, 2008, pp. 88--96.

\bibitem{qin2010general}
T.~Qin, T.-Y. Liu, and H.~Li, ``A general approximation framework for direct
  optimization of information retrieval measures,'' \emph{Information
  Retrieval}, vol.~13, no.~4, pp. 375--397, 2010.

\bibitem{bruch2019revisiting}
S.~Bruch, M.~Zoghi, M.~Bendersky, and M.~Najork, ``Revisiting approximate
  metric optimization in the age of deep neural networks,'' in \emph{Proc. Int.
  ACM SIGIR Conf. Res. Develop. Inf. Retr.~({SIGIR})}, 2019, pp. 1241--1244.

\bibitem{reimers2019sentence}
N.~Reimers and I.~Gurevych, ``Sentence-bert: Sentence embeddings using siamese
  bert-networks,'' in \emph{Proc. Conf. Empir. Meth. Natural Lang.
  Process.~({EMNLP})}, 2019, pp. 3982--3992.

\bibitem{young2014image}
P.~Young, A.~Lai, M.~Hodosh, and J.~Hockenmaier, ``From image descriptions to
  visual denotations: New similarity metrics for semantic inference over event
  descriptions,'' \emph{Trans. Assoc. Comput. Linguist.}, vol.~2, pp. 67--78,
  2014.

\bibitem{lin2014microsoft}
T.-Y. Lin, M.~Maire, S.~Belongie, J.~Hays, P.~Perona, D.~Ramanan,
  P.~Doll{\'a}r, and C.~L. Zitnick, ``Microsoft coco: Common objects in
  context,'' in \emph{Proc. Eur. Conf. Comput. Vis.~({ECCV})}.\hskip 1em plus
  0.5em minus 0.4em\relax Springer, 2014, pp. 740--755.

\bibitem{wang2019camp}
Z.~Wang, X.~Liu, H.~Li, L.~Sheng, J.~Yan, X.~Wang, and J.~Shao, ``Camp:
  Cross-modal adaptive message passing for text-image retrieval,'' in
  \emph{Proc. {IEEE/CVF} Int. Conf. Comput. Vis.~({ICCV})}, 2019, pp.
  5764--5773.

\bibitem{liu2019focus}
C.~Liu, Z.~Mao, A.-A. Liu, T.~Zhang, B.~Wang, and Y.~Zhang, ``Focus your
  attention: A bidirectional focal attention network for image-text matching,''
  in \emph{Proc. ACM Int. Conf. Multimedia~({ACM MM})}, 2019, pp. 3--11.

\bibitem{wang2019position}
Y.~Wang, H.~Yang, X.~Qian, L.~Ma, J.~Lu, B.~Li, and X.~Fan, ``Position focused
  attention network for image-text matching,'' in \emph{Proc. Int. Joint Conf.
  Artif. Intell.~({IJCAI})}, 2019, pp. 3792--3798.

\bibitem{wang2020pfan++}
Y.~Wang, H.~Yang, X.~Bai, X.~Qian, L.~Ma, J.~Lu, B.~Li, and X.~Fan, ``Pfan++:
  Bi-directional image-text retrieval with position focused attention
  network,'' \emph{{IEEE} Trans. Multimedia}, vol.~23, pp. 3362--3376, 2020.

\bibitem{wu2021region}
J.~Wu, C.~Wu, J.~Lu, L.~Wang, and X.~Cui, ``Region reinforcement network with
  topic constraint for image-text matching,'' \emph{{IEEE} Trans. Circuits
  Syst. Video Technol.}, vol.~32, no.~1, pp. 388--397, 2022.

\bibitem{wei2020multi}
X.~Wei, T.~Zhang, Y.~Li, Y.~Zhang, and F.~Wu, ``Multi-modality cross attention
  network for image and sentence matching,'' in \emph{Proc. {IEEE/CVF} Conf.
  Comput. Vis. and Pattern Recognit.~({CVPR})}, 2020, pp. 10\,941--10\,950.

\bibitem{zhang2022unified}
K.~Zhang, Z.~Mao, A.~Liu, and Y.~Zhang, ``Unified adaptive relevance
  distinguishable attention network for image-text matching,'' \emph{{IEEE}
  Trans. Multimedia}, pp. 1--1, 2022.

\bibitem{yang2022dual}
S.~Yang, Q.~Li, W.~Li, X.~Li, and A.-A. Liu, ``Dual-level representation
  enhancement on characteristic and context for image-text retrieval,''
  \emph{{IEEE} Trans. Circuits Syst. Video Technol.}, vol.~32, no.~11, pp.
  8037--8050, 2022.

\bibitem{zhang2022show}
H.~Zhang, Z.~Mao, K.~Zhang, and Y.~Zhang, ``Show your faith: Cross-modal
  confidence-aware network for image-text matching,'' in \emph{Proc. AAAI Conf.
  Artif. Intell.~({AAAI})}, vol.~36, no.~3, 2022, pp. 3262--3270.

\bibitem{wang2022dual}
Y.~Wang, Y.~Su, W.~Li, J.~Xiao, X.~Li, and A.-A. Liu, ``Dual-path rare content
  enhancement network for image and text matching,'' \emph{{IEEE} Trans.
  Circuits Syst. Video Technol.}, pp. 1--1, 2023.

\bibitem{li2022image}
K.~Li, Y.~Zhang, K.~Li, Y.~Li, and Y.~Fu, ``Image-text embedding learning via
  visual and textual semantic reasoning,'' \emph{{IEEE} Trans. Pattern Anal.
  Mach. Intell.}, 2022.

\bibitem{wen2021learning}
K.~Wen, X.~Gu, and Q.~Cheng, ``Learning dual semantic relations with graph
  attention for image-text matching,'' \emph{{IEEE} Trans. Circuits Syst. Video
  Technol.}, vol.~31, no.~7, pp. 2866--2879, 2021.

\bibitem{anderson2018bottom}
P.~Anderson, X.~He, C.~Buehler, D.~Teney, M.~Johnson, S.~Gould, and L.~Zhang,
  ``Bottom-up and top-down attention for image captioning and visual question
  answering,'' in \emph{Proc. {IEEE/CVF} Conf. Comput. Vis. and Pattern
  Recognit.~({CVPR})}, 2018, pp. 6077--6086.

\bibitem{he2016deep}
K.~He, X.~Zhang, S.~Ren, and J.~Sun, ``Deep residual learning for image
  recognition,'' in \emph{Proc. {IEEE/CVF} Conf. Comput. Vis. and Pattern
  Recognit.~({CVPR})}, 2016, pp. 770--778.

\bibitem{kingma2015adam}
D.~P. Kingma and J.~Ba, ``Adam: A method for stochastic optimization,'' in
  \emph{Proc. Int. Conf. Learn. Represent.~({ICLR})}, 2015.

\bibitem{loshchilov2018decoupled}
I.~Loshchilov and F.~Hutter, ``Decoupled weight decay regularization,'' in
  \emph{Proc. Int. Conf. Learn. Represent.~({ICLR})}, 2019.

\bibitem{xuan2020hard}
H.~Xuan, A.~Stylianou, X.~Liu, and R.~Pless, ``Hard negative examples are hard,
  but useful,'' in \emph{Proc. Eur. Conf. Comput. Vis.~({ECCV})}, 2020, pp.
  126--142.

\end{thebibliography}
\bibliographystyle{IEEEtran}

\end{document}